\documentclass[11pt]{article}
\usepackage[margin=1.15in]{geometry}
\usepackage[T1]{fontenc}
\usepackage{amsmath,amssymb,amsthm}
\usepackage{natbib}
\usepackage{hyperref}
\usepackage{booktabs}
\usepackage{tikz}
\theoremstyle{plain}
\newtheorem{theorem}{Theorem}
\newtheorem{proposition}[theorem]{Proposition}
\newtheorem{corollary}[theorem]{Corollary}
\newtheorem{lemma}[theorem]{Lemma}
\theoremstyle{definition}
\newtheorem{definition}[theorem]{Definition}
\newtheorem{axiom}[theorem]{Axiom}
\newtheorem{assumption}[theorem]{Assumption}
\newtheorem{remark}[theorem]{Remark}

\newcommand{\Sti}{\Sigma}
\newcommand{\parti}[1]{\Pi_{#1}}
\newcommand{\finer}{\preceq}

\title{What Lies Beyond Perception:\\ A Partition Theory of the Limits\\ of AI Evaluation and Oversight}
\author{Ahmed M. Mahran\\
Incorta\\
\texttt{ahmed.mahran@incorta.com} \quad \texttt{ahmahran@gmail.com}}
\date{October 2026}

\begin{document}
\maketitle

\begin{abstract}
Replacing the judge can move a benchmark score, and models can be trained to hide capabilities. Both put the evaluator in question, not only the model: what can it certify? We develop a partition theory of evaluation from one principle: judgement is bounded by perception. An agent's discrimination is the partition of its inputs into classes it cannot tell apart, and its outputs depend on an input only through that class. Discrimination is necessary for capability but not sufficient: a model cannot perform a task that requires a distinction it does not draw, and drawing the distinction does not ensure the task is done. Bounds on discrimination therefore bear directly on the question that matters most for safety: whether evaluation can show that a model lacks an ability. We prove a one-sided bound. An examinee whose partition refines its evaluator's has the discrimination to pass every test the evaluator can grade, provided each input has an acceptable reply. We define the echo test, which sends an input and accepts any reply the evaluator perceives as identical to it. The echo test is complete for the tests the evaluator can grade, much as a complete problem stands for its complexity class: no test the evaluator can grade certifies more about an examinee's discrimination. An examinee that conflates two inputs the evaluator separates fails it once both inputs are probed. Passing therefore certifies at most a floor on discrimination, never a ceiling. Further elicitation of the examinee cannot raise this floor; only a sharper evaluator can. Used to detect concealment, echo-test failure points the wrong way: under exhaustive probing, it flags every honestly weaker examinee and no concealer that presents as the evaluator's peer. Exact imitation of a weaker model is undetectable by any behavioural evaluator. We extend the bound to several observers. An added monitor sharpens the evaluation only if it separates inputs that no existing monitor separates. A model that controls one channel, such as its reasoning trace, can hide only what the channels it cannot influence fail to separate, provided the evaluator checks the channels against one another. We use the theory to explain published findings on judge disagreement and sandbagging, and we derive predictions that experiments can test. Behavioural evaluation can establish that a model draws a distinction. Without the model's cooperation, it cannot establish that the model lacks one. A safety case that argues from behavioural evaluation alone that a model lacks an ability must therefore assume that cooperation.
\end{abstract}

\section{Introduction}\label{sec:intro}

It is by now a commonplace of AI evaluation that tests bound capability from below only. \citet{barnett2024} state it directly: evaluations ``can establish lower bounds on AI
capabilities'' while facing ``an inability to establish upper bounds on capabilities.'' \citet{vanderweij2024} report the practical form: models can be fine-tuned to
``approximately emulate a consistent capability profile of a weaker model.''

The usual explanation is \emph{under-elicitation}: a model may possess a capability that
prompting, scaffolding or fine-tuning failed to surface, so a negative result does not establish
absence, and a claim of the form ``this model cannot do $X$'' is only as good as the effort spent
asking. That explanation concerns the examinee. Evaluations also depend on the evaluator: a benchmark score can move when the judge is replaced and the candidate responses are held fixed
\citep{judgechange2026}, and a judge scores a candidate higher when that candidate's mistakes
resemble its own \citep{capa2025}.

This paper studies the evaluator. Its thesis is that judgement is bounded by perception: what an evaluator can certify is bounded by its resolution, the distinctions it can draw, under its current elicitation (Lemma~\ref{lem:bound} and Corollary~\ref{cor:asym}). \citet{mahran2012} took the principle that judgement depends on perception, with a perception--action model of agents, to the question of whether agents can judge that they perceive alike, and concluded that a judge comparing percepts can certify a difference but not a sameness. Here that line is developed formally, as a partition theory, for the evaluation of AI systems. Model an agent by
the partition its perception induces on the space of inputs: two inputs are equivalent for that
agent when it cannot tell them apart. Subject to one axiom, that an agent's output depends on its input only through its percept, two facts follow.

Suppose the examinee separates every pair of inputs its evaluator separates, and perhaps more. Then, if it knows how the evaluator perceives, it can work out which inputs the evaluator treats as the same, and answer as though it made no
finer distinctions itself. Its extra resolution leaves no trace, and nothing in what the evaluator
sees sets it apart from an agent that genuinely has no more.

Now suppose the opposite: some pair of inputs looks identical to the examinee and different to the
evaluator. The examinee must answer both the same way, because it cannot see that they differ. The
evaluator can, so it sees one answer where it expected two. That is a failure the examinee cannot
avoid.

The two are not symmetric, and the reason is that discarding distinctions is easy and inventing
them is not. An agent that sees finely can always group its percepts and act as though it saw
less. An agent that sees coarsely cannot act as though it saw more: on any pair it has already
conflated, its response is fixed. So discrimination above the evaluator's resolution can be hidden and a shortfall below it is exposed, and exhaustive probing locates a floor while leaving everything above it open. We call the two emulation from above and detection from below (Theorems~\ref{thm:emulate} and~\ref{thm:detect}); together they give the one-sided bound.

The evaluator is not exempt. It is itself an agent, with its own limit on what it can tell apart,
so what it establishes is bounded by that limit, and certifying the limit would take a sharper
evaluator facing the same problem. Evaluations are run by agents, whether language models or
humans, so this regress has no privileged end point.

The most an evaluation can certify is therefore set by the evaluator's resolution. Eliciting the examinee harder does not
raise it, since a sharper examinee can reproduce whatever its evaluator expects however hard it is asked; eliciting the evaluator can.

A sharper evaluator is the obvious remedy, and a real one. Three things about it are not
obvious. The first is how sharp. The evaluator must separate the concealer's reply from the reply of the agent it presents as, at some
stimulus it probes (\S\ref{sub:sandbag}), a necessary condition, not a sufficient one (Remark~\ref{rem:model}), and not guaranteed by being merely sharper than the last evaluator. Behavioural effort spent on the examinee does not expose an exact copy, and neither does effort on the evaluator (Corollary~\ref{cor:sandbag}); against inexact concealment, effort on the evaluator helps only up to the sharpest evaluator available; for a system at the frontier a single
sharp enough evaluator may not exist. An ensemble offers a way
round this. Its join separates every pair that some member separates (Definition~\ref{def:join}), so a collection can be sharper than every one of its members, each supplying distinctions the others lack. The route to a sharper instrument can therefore run through diversity of discrimination as well as through a better individual judge. The second is that no warning is
issued. By Corollary~\ref{cor:asym} passage says only that the examinee is somewhere above the
evaluator, so a pass from an evaluator too coarse for the task looks exactly like a pass from an adequate one. The third is that the usual instruments do not act on this:
harder items and richer scaffolds press on the examinee, and change nothing about what the
instrument doing the reading can tell apart.

Emulation from above and detection from below both concern an evaluator reading behaviour: emulation requires the examinee to choose
what the evaluator sees, and detection requires a reply the evaluator can perceive. The part of Corollary~\ref{cor:asym} that needs no strategy on the examinee's side, item 2, carries to an observer of any kind: an observation within a cell is consistent with everything in it. \S\ref{sub:absence} uses this for interpretability probes.

\paragraph{Contributions.}
\begin{enumerate}\itemsep2pt
\item \textbf{A framing} (Sections~\ref{sec:model} and~\ref{sec:main}): the evaluator, not only the
examinee, has a discrimination structure, and what it can certify is bounded by it
(Lemma~\ref{lem:bound}). An evaluation compares two such structures, which gives a one-sided bound:
an examinee sharper than its evaluator can present as a peer, while one that is coarser cannot conceal the fact from exhaustive probing. The bound holds for tests the evaluator can grade under the conditions of Theorems~\ref{thm:emulate} and~\ref{thm:detect}; the echo test is complete for such tests (Proposition~\ref{prop:finest}); a coarse evaluator errs against honest weakness (Corollary~\ref{cor:spec}); and a worked example instantiates it (Appendix~\ref{sub:example}).
\item \textbf{Six consequences} (Section~\ref{sec:cons}), arranged on a $2\times3$ grid of who assesses whom against adversariality. Exact sandbagging is undetectable by any behavioural evaluator. An interpretability probe cannot prove absence where it merges representations. An added monitor increases collective discrimination exactly when it separates a pair the ensemble does not. A channel the examinee controls can hide, from an evaluator that checks consistency over undisturbed channels, only what every other channel misses. No discrimination ceiling is self-assessed, and adversarial self-report adds nothing to testing.
Corollary~\ref{cor:necessity} limits these results to discrimination, which is necessary for task competence but not sufficient.
\item \textbf{Empirical implications} (Section~\ref{sec:predictions}). Four published findings read
through the model: judges disagree, weight noise exposes sandbagging, elicitation recovers locked
capability, and monitor-aware models sometimes evade chain-of-thought monitors. Three predictions follow. For concealment that the examinee's elicitation leaves in place, eliciting the examinee should not raise detection, while sharpening the evaluator can. Under weight noise a concealing model returns to its unconcealed replies, with a rise that shrinks as the task needs more layers of processing. Encodings shaped against a monitor, over inputs that vary enough to isolate each output, approach renamings within the level sets of its score.
\end{enumerate}

\section{Model}\label{sec:model}

Let $\Sti$ be a set of stimuli. An element $s \in \Sti$ is a single thing an agent may be
presented with: a prompt for a language model, a situation for an embodied agent. An agent $A$ has a percept space $X_A$ and a
perception map $P_A : \Sti \to X_A$, and emits actions from a set $\mathcal{A}$ via $E_A : X_A \to \mathcal{A}$.

\begin{definition}[Discrimination partition]\label{def:partition}
$P_A$ induces the equivalence $s \equiv_A s' \iff P_A(s) = P_A(s')$. Write $\parti{A}$ for the
resulting partition of $\Sti$. We say $\parti{B} \finer \parti{A}$ ($B$ is \emph{at least as sharp}
as $A$) when $\parti{B}$ refines $\parti{A}$: every $B$-cell lies within an $A$-cell, i.e.\
$s \equiv_B s' \implies s \equiv_A s'$.
\end{definition}

\begin{definition}[Join]\label{def:join}
The \emph{join} $\parti{1} \vee \parti{2}$ of two partitions is the coarsest partition refining
both: two stimuli share a cell of the join exactly when they share a cell of each. For a family, $\bigvee_i \parti{i}$ puts two stimuli together exactly when every $\parti{i}$ does.\footnote{The name follows Figure~\ref{fig:lattice}, which draws finer partitions higher. Under $\finer$, which puts finer partitions lower, the same partition is the meet.}
\end{definition}

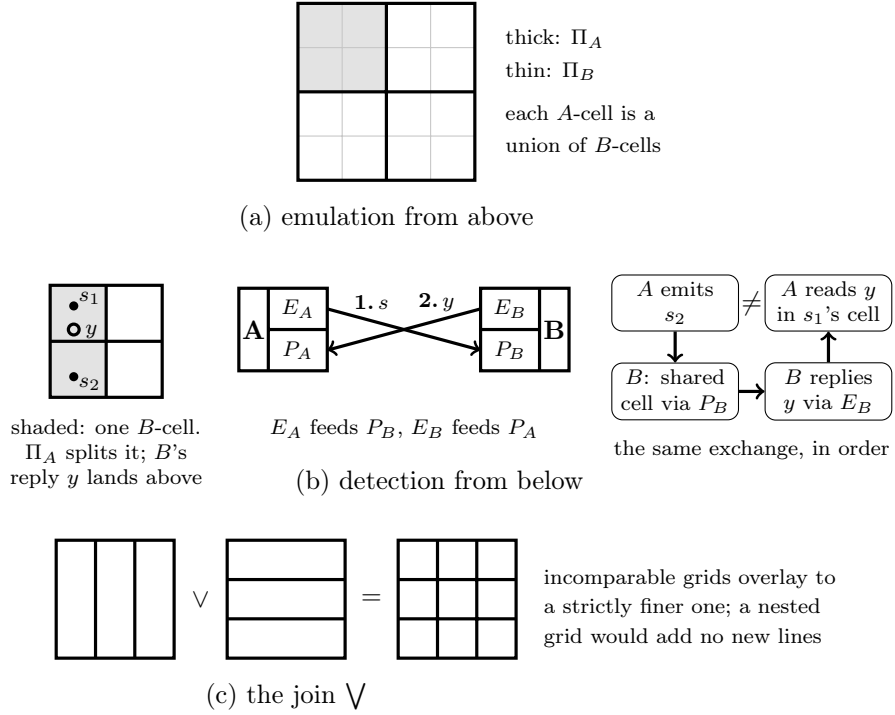
\begin{figure}[htbp]
\centering
%%%%%%%%%%%%%%%%%%%%%%%% (a) emulation from above
\begin{tikzpicture}[scale=0.78, every node/.style={font=\small}]
 \fill[gray!22] (0,1.5) rectangle (1.5,3);
 \draw[gray!45, thin] (0.75,0)--(0.75,3) (2.25,0)--(2.25,3)
 (0,0.75)--(3,0.75) (0,2.25)--(3,2.25);
 \draw[very thick] (0,0) rectangle (3,3);
 \draw[very thick] (1.5,0)--(1.5,3) (0,1.5)--(3,1.5);
 \node[anchor=west, font=\scriptsize] at (3.35,2.4) {thick: $\parti{A}$};
 \node[anchor=west, font=\scriptsize] at (3.35,1.85) {thin: $\parti{B}$};
 \node[anchor=west, font=\scriptsize] at (3.35,1.15) {each $A$-cell is a};
 \node[anchor=west, font=\scriptsize] at (3.35,0.6) {union of $B$-cells};
 \node at (1.5,-0.65) {(a) emulation from above};
\end{tikzpicture}

\vspace{4mm}

%%%%%%%%%%%%%%%%%%%%%%%% (b) detection from below -- conflation, composition, dialogue
\begin{tikzpicture}[scale=0.78, every node/.style={font=\small}]
 % the conflation: shading is B's single cell; every line of A is thick, including
 % the one B lacks -- that line is the whole content of the panel.
 \fill[gray!22] (0,0) rectangle (0.95,1.9);
 \draw[very thick] (0,0) rectangle (1.9,1.9);
 \draw[very thick] (0.95,0)--(0.95,1.9);
 \draw[very thick] (0,0.95)--(1.9,0.95);
 \fill (0.40,1.55) circle (2.2pt); \node[font=\scriptsize] at (0.67,1.68) {$s_1$};
 \fill (0.40,0.35) circle (2.2pt); \node[font=\scriptsize] at (0.67,0.22) {$s_2$};
 \draw[very thick] (0.40,1.15) circle (2.6pt);
 \node[font=\scriptsize] at (0.70,1.12) {$y$};
 \node[anchor=north, align=center, font=\scriptsize] at (0.95,-0.20)
 {shaded: one $B$-cell.\\$\parti{A}$ splits it; $B$'s\\reply $y$ lands above};

 % the composition, as a crossing
 \draw[very thick] (3.2,0.45) rectangle (4.7,1.85);
 \draw[very thick] (3.7,0.45)--(3.7,1.85) (3.7,1.15)--(4.7,1.15);
 \node at (3.45,1.15) {\bfseries A};
 \node[font=\scriptsize] at (4.2,1.50) {$E_A$}; \node[font=\scriptsize] at (4.2,0.80) {$P_A$};
 \draw[very thick] (7.3,0.45) rectangle (8.8,1.85);
 \draw[very thick] (8.3,0.45)--(8.3,1.85) (7.3,1.15)--(8.3,1.15);
 \node at (8.55,1.15) {\bfseries B};
 \node[font=\scriptsize] at (7.8,1.50) {$E_B$}; \node[font=\scriptsize] at (7.8,0.80) {$P_B$};
 \draw[->, very thick] (4.7,1.50) -- (7.3,0.80);
 \draw[->, very thick] (7.3,1.50) -- (4.7,0.80);
 \node[font=\scriptsize] at (5.45,1.62) {\textbf{1.}\,$s$};
 \node[font=\scriptsize] at (6.55,1.62) {\textbf{2.}\,$y$};
 \node[anchor=north, align=center, font=\scriptsize] at (6.0,-0.20)
 {$E_A$ feeds $P_B$, $E_B$ feeds $P_A$};

 % the dialogue, as a U
 \tikzset{stepbox/.style={draw, rounded corners, font=\scriptsize, align=center,
 inner sep=2.5pt, text width=1.5cm}}
 \node[stepbox] (e1) at (10.6,1.60) {$A$ emits\\$s_2$};
 \node[stepbox] (e2) at (10.6,0.10) {$B$: shared\\cell via $P_B$};
 \node[stepbox] (e3) at (13.2,0.10) {$B$ replies\\$y$ via $E_B$};
 \node[stepbox] (e4) at (13.2,1.60) {$A$ reads $y$\\in $s_1$'s cell};
 \draw[->, very thick] (e1)--(e2);
 \draw[->, very thick] (e2)--(e3);
 \draw[->, very thick] (e3)--(e4);
 \node[font=\small] at (11.9,1.60) {$\neq$};
 \node[anchor=north, align=center, font=\scriptsize] at (11.9,-0.55)
 {the same exchange, in order};
 \node at (6.6,-1.45) {(b) detection from below};
\end{tikzpicture}

\vspace{4mm}

%%%%%%%%%%%%%%%%%%%%%%%% (c) the join
\begin{tikzpicture}[scale=0.78, every node/.style={font=\small}]
 \draw[very thick] (0,0) rectangle (2,2);
 \draw[very thick] (0.667,0)--(0.667,2) (1.333,0)--(1.333,2);
 \node at (2.45,1) {$\vee$};
 \draw[very thick] (2.9,0) rectangle (4.9,2);
 \draw[very thick] (2.9,0.667)--(4.9,0.667) (2.9,1.333)--(4.9,1.333);
 \node at (5.35,1) {$=$};
 \draw[very thick] (5.8,0) rectangle (7.8,2);
 \draw[very thick] (6.467,0)--(6.467,2) (7.133,0)--(7.133,2)
 (5.8,0.667)--(7.8,0.667) (5.8,1.333)--(7.8,1.333);
 \node[anchor=west, font=\scriptsize] at (8.1,1.35) {incomparable grids overlay to};
 \node[anchor=west, font=\scriptsize] at (8.1,0.85) {a strictly finer one; a nested};
 \node[anchor=west, font=\scriptsize] at (8.1,0.35) {grid would add no new lines};
 \node at (3.9,-0.65) {(c) the join $\bigvee$};
\end{tikzpicture}
\caption{The three results as grids ruled over the stimulus space $\Sti$. $A$ is the evaluator
and its lines are always thick; $B$ is the examinee. Shading marks one cell of whichever agent is
coarser: $A$ in (a), $B$ in (b). \textbf{(a)} Every $A$-cell, such as the shaded one, is exactly a union of $B$-cells, so $B$ can always name the $A$-cell it is in and report as a peer
(Theorem~\ref{thm:emulate}). \textbf{(b)} $B$'s grid lacks the horizontal line, so the shaded
column is one $B$-cell while $\parti{A}$ splits it, and $s_1$, $s_2$ are indistinguishable to $B$
but distinct to $A$. $B$'s reply $y$ (hollow) represents that whole cell and lies in $s_1$'s
$A$-cell, so when sent $s_2$, $A$ receives something it reads as $s_1$
(Theorem~\ref{thm:detect}). The crossing is the composition $P_A E_B P_B E_A$, and the
$\mathrm{U}$ beside it lays out the same exchange in order. \textbf{(c)} Collective
discrimination is the overlay of grids, and only lines not already present add cells
(Proposition~\ref{prop:join}). Panels (a) and (b) draw comparable partitions; incomparable ones
overlay as in (c).}
\label{fig:grids}
\end{figure}

A partition can be read as a grid ruled over $\Sti$: the agent sees which cell a stimulus
falls into, and nothing finer. Refinement is a grid with more lines, and the join of two grids is their overlay. Figure~\ref{fig:grids} renders the results of Sections~\ref{sec:main}
and~\ref{sec:cons} this way.

The grid picture can obscure two properties of partitions. Not every partition is a grid; cells
may take any shape.
And refinement is a partial order rather than a total one, so two partitions need not be comparable: a cell of one may straddle a boundary of the other, each separating what the other
runs together. Nested grids are therefore not guaranteed, and we expect them to be the exception rather than the rule. The theorems split on whether $\parti{B}$ refines $\parti{A}$, and the incomparable case falls on the side where it does not.

\paragraph{Notation.} Agents are $A, B, C, \dots$. A letter names an agent rather than a place in
a protocol: in the statements, $A$ is the evaluator and $B$ the examinee, whether the setting is the echo test of Definition~\ref{def:echo}, the interpretability probe of \S\ref{sub:absence} or the language model judge of a scalable-oversight pipeline. A proof that applies a result with another agent in $A$'s role says so. For an agent $Z$ we write $P_Z$ for its perception, $E_Z$ for its emission, $\beta_Z = E_Z \circ P_Z$ for its behaviour and $\parti{Z}$ for its discrimination partition; a \emph{peer} of $A$ is an agent whose partition is $\parti{A}$. Lowercase indices $\parti{i}, \parti{j}$ denote observers or channels rather than named agents. One symbol
stands outside this scheme: $J$, a percept-reading evaluator that observes percepts directly rather than behaviour,
which appears only where we contrast the two kinds of access (Figure~\ref{fig:access}, Appendix~\ref{app:s3}). Table~\ref{tab:notation} in Appendix~\ref{app:notation} collects the symbols.

In a neural network the partition has a mechanism behind it. A network represents more features than it has
dimensions by storing them in overlapping directions and tolerating the interference, a regime
known as superposition \citep{elhage2022}. Capacity is finite and features compete for it, so a
model separates what it has room to separate. Features it drops from a representation are run together there exactly, which is a partition, unless another route, such as copying, carries them; features stored in superposition are separated only up to interference, a graded case Section~\ref{sec:limits} returns to.

\paragraph{Where the boundary falls.} On this view a partition is fixed by construction, and what
counts as a change of partition depends on where the boundary of the agent is drawn. For a language model there are two natural boundaries. Take the agent to be the bare model, and $\parti{A}$ is a property of its weights. A prompt, a scaffold or a
returned tool result is then a further input rather than a modification of the thing receiving
it, so it changes the stimulus and not the partition, while fine-tuning changes the weights and so
yields a different agent. Take the agent to be the deployed system, meaning the model together
with its prompt template, scaffold, tools and decoding procedure, and $\parti{A}$ is a property of
that whole system; changing any of these yields a different system, which may differ in its
partition, its emission or both. Which boundary to use for each party is the study's choice; the theory compares any two partitions of the same stimuli, wherever their boundaries are drawn. Assumption~\ref{asm:monotone} (Section~\ref{sec:predictions}) is accordingly a claim about the configurations
an elicitation budget produces, not a claim that perception itself improves.

We do not model the medium, meaning anything that acts on a stimulus between the two agents and is
controlled by neither, whether deterministic or random. A reply is taken to arrive unchanged. This matters for the echo test, in which $A$ reads $B$'s reply as a stimulus
(Definition~\ref{def:echo}).
A study that needs a medium can add it as a further map on stimuli between the two agents; this
paper does not.

\begin{axiom}[Perception mediates behaviour]\label{ax:noshortcut}
An agent's behaviour depends on the stimulus only through its percept: the total behaviour map
$\beta_A : \Sti \to \mathcal{A}$ factors as $\beta_A = E_A \circ P_A$.
\end{axiom}

Axiom~\ref{ax:noshortcut} makes perception the single map behind everything an agent does: elicitation can change the percept or how it becomes a reply, but every reply is computed from the percept. The axiom does not say how perception and emission are implemented. In a language model, for instance, the same weights may serve both, and decoding, which acts after the forward pass, belongs to emission: greedy decoding gives a map, and sampling brings in the seed discussed below. Every route by which the stimulus reaches the reply counts toward the partition, including verbatim copying and lookup in a fixed cache or retrieval index (a cache written between probes is state, which the axiom excludes). A system that can return an input verbatim separates that input from every other, so its partition is fine whatever it understands. Such a partition says nothing about competence, since discrimination is necessary for competence and not sufficient (Corollary~\ref{cor:necessity}). The results take an agent's emission to range over all maps from its percepts to actions: that $B$ can realise a behaviour means some emission gives it, and whether $B$ can compute that emission is a further capability the model does not represent. Section~\ref{sec:limits} returns to this.

\begin{lemma}[Judgement is bounded by perception]\label{lem:bound}
Anything an agent computes from its percepts is constant on the cells of its partition: if
$f = g \circ P_A$ for some map $g$, then $s \equiv_A s'$ implies $f(s) = f(s')$. In particular,
by Axiom~\ref{ax:noshortcut}, $\beta_A$ is constant on the cells of $\parti{A}$.
\end{lemma}

\begin{proof}
If $s \equiv_A s'$ then $P_A(s) = P_A(s')$, so $g(P_A(s)) = g(P_A(s'))$.
\end{proof}

A verdict, a self-report and a monitor's flag are all computed from percepts, so each is bounded in
this way. 

A behaviour $\beta : \Sti \to \mathcal{A}$
\emph{factors through} a perception map $P : \Sti \to X$ when there is an emission map
$E : X \to \mathcal{A}$ with $\beta = E \circ P$; equivalently, $\beta$ is constant on the cells
of the partition $P$ induces, so it separates no pair that $P$ conflates.
Axiom~\ref{ax:noshortcut} says every agent's behaviour factors through its own perception map.
Theorem~\ref{thm:emulate} concerns behaviour that factors through another agent's.

The axiom says two things. First, behaviour is a function of the stimulus: the same stimulus always gets the same reply, so no state is carried from one probe to the next. Second, that function passes through the percept: it factors through $P_A$. Determinism therefore follows from the first part and need not be assumed separately. An agent whose reply varies while its percept stays fixed must be reading something besides that percept. It may be state, which the axiom excludes. Or it may be a random seed, such as the one a sampler draws to pick each reply. A seed is an input, so we count it as part of the stimulus: two probes with different seeds are then different stimuli, and $\beta_A$ is again a function, on a stimulus set enlarged to include seeds. For a language model with greedy decoding there is no seed, and the stimulus set needs no enlarging.

\begin{definition}[Tests]\label{def:test}
Fix an examinee $B$.
\begin{enumerate}\itemsep1pt
\item A \emph{test} $\mathcal{T} = (S, G)$ consists of a set $S \subseteq \Sti$ of \emph{test cases} and an \emph{acceptance rule} $G$ that assigns to each $s \in S$ a set $G(s) \subseteq \mathcal{A}$ of acceptable replies.
\item $B$ \emph{passes} $\mathcal{T}$ at $s \in S$ when $\beta_B(s) \in G(s)$.
\item An evaluator administers $\mathcal{T}$ by probing a set $Q \subseteq S$ of its cases; each presentation of a case is a \emph{probe}, and a case may be probed more than once. The evaluator probes \emph{exhaustively} when $Q = S$. $B$ \emph{passes} $\mathcal{T}$ on $Q$ when it passes at every $s \in Q$.
\item $\mathcal{T}$ is \emph{gradable by} an evaluator $A$ when its acceptance rule factors through $A$'s perception: for $s, s' \in S$, $s \equiv_A s'$ implies $G(s) = G(s')$. $\mathcal{T}$ is \emph{graded by} $A$ when $A$ is the evaluator that administers it.
\item A \emph{task} is a test whose acceptable replies are the correct responses.
\end{enumerate}
\end{definition}

A test fixes its cases and its rule, whoever supplies them. A claim about all of $\Sti$ needs $S = \Sti$; a claim about a narrower scope takes that scope as $S$. A test gradable by $A$ is one whose acceptable replies are fixed by what $A$ perceives of each case. A benchmark scored by a judge against reference answers is graded by that judge, and gradable by it when the judge tells apart any two items whose reference answers differ. A task need not be gradable by $A$. When $A$ grades a test by perceiving each case and each reply separately, its verdicts depend only on those percepts (Lemma~\ref{lem:bound}), so it applies some rule gradable by $A$. If the test is not gradable by $A$, that rule differs from the test's own at one of two cases $A$ conflates, and there $A$ accepts or rejects some reply wrongly.

The results below take the cases of a test to be all of $\Sti$, the setting of a global claim. Each holds for a narrower scope $S$ with the cases restricted to $S$ and every partition restricted to $S$, that is, replaced by the partition of $S$ into the non-empty intersections of its cells with $S$. Replies are still perceived through $P_A$ on all of $\Sti$, and a condition such as ``every $A$-cell contains an action'' then concerns the $A$-cells that meet $S$. A pass then certifies refinement on $S$ only.

\begin{definition}[Echo test]\label{def:echo}
Assume the channel is \emph{closed}: $\mathcal{A} \subseteq \Sti$, so that a reply is itself a
stimulus $A$ can perceive. (For sequence models, whose prompts and replies share a stimulus space,
this is immediate. Where a reply is instead scored outside the channel, by executing it, by
inspecting side effects or by a human marker, the containment $\mathcal{A} \subseteq \Sti$ has to be
assumed rather than inherited from the setup.) The \emph{echo test} of $A$ is the test with cases $\Sti$, gradable by $A$ by construction, with acceptance rule
\[
  G(s) \;=\; \{\, y \in \mathcal{A} : P_A(y) = P_A(s) \,\}.
\]
$A$ sends $s$, and $B$ passes at $s$ when $A$ receives back something it cannot distinguish from
what it sent.
\end{definition}

A system that can copy its input passes the echo test trivially. Copying separates every input, so the copier's partition is fine, and its pass certifies that discrimination and nothing about competence (Corollary~\ref{cor:necessity}). A test meant to measure competence rather than discrimination can ask for a paraphrase instead.

The echo test asks only whether $B$ can return what $A$ said, as judged by $A$, so $A$'s partition
is the only standard in play. It does not ask $B$ to return $s$ itself, only a reply in the same
$A$-cell.  In the echo test, every two stimuli $A$ separates have disjoint sets of acceptable replies.

\section{The asymmetry}\label{sec:main}

Two facts, pointing opposite ways. An examinee at least as sharp as its evaluator can reproduce
whatever that evaluator expects. One that is coarser cannot hide the fact from an evaluator that probes a pair it conflates. These are emulation from above and detection from below, named after the examinee's place relative to its evaluator. Both follow from Lemma~\ref{lem:bound}, and together they give the one-sided bound. They hold for tests the evaluator can grade, under conditions on the acceptable replies.

\begin{theorem}[Emulation from above]\label{thm:emulate}
Let $\parti{B} \finer \parti{A}$.
\begin{enumerate}\itemsep1pt
\item $B$ can realise the behaviour of \emph{any} agent whose behaviour depends on the stimulus
only through $P_A$, and is then accepted by every test exactly where that agent is.
\item $B$ can pass, at every stimulus, any test gradable by $A$ in which every set of acceptable
replies is non-empty. In particular, if every $A$-cell contains an action, $B$ can pass the echo test at every stimulus and be behaviourally indistinguishable from a peer of $A$.
\end{enumerate}
\end{theorem}

The proof builds the emulator from two maps. Since every $B$-cell lies inside one $A$-cell, $B$ can tell which $A$-cell an input is in; the \emph{coarsening map} $h$ does this, turning $B$'s percept into $A$'s. If every $A$-cell contains an action, the \emph{prototype selector} $\rho$ picks one action in each $A$-cell. With $E_B = \rho \circ h$, $B$ answers each input with an action from that input's $A$-cell, which $A$ cannot tell from the input itself. The proof is in Appendix~\ref{app:s3}.

Theorem~\ref{thm:emulate} says such an emission exists, not that it is easy to find or compute: in item 1, $B$'s emission must compute the target's whole emission $E$ together with the coarsening $h$. Whether $B$ can compute that emission is a further capability, which the model does not represent. Remark~\ref{rem:model} names one part of it.

\begin{remark}[The emulator must model the evaluator]\label{rem:model}
Theorem~\ref{thm:emulate} makes emulation possible. To emulate a particular target, $B$ must also know how that target behaves, as $A$ perceives it. If $B$ misjudges its target, it is not thereby exposed: it simply behaves differently from the target. No behaviour is evidence of concealment on its own. Any behaviour $\beta_B$ is exactly that of an agent that conceals nothing, namely the agent whose percept is $\beta_B$ itself and whose emission is the identity. Behaviour reveals concealment only relative to an assumption about what honest agents look like, for example that an honest weak model makes certain mistakes. Without such an assumption, an emulator that misjudges its target looks like some other honest agent.
\end{remark}

\begin{theorem}[Detection from below]\label{thm:detect}
Let $\mathcal{T} = (S, G)$ be any test, and let $s_1, s_2 \in S$ with $s_1 \equiv_B s_2$ and $G(s_1) \cap G(s_2) = \emptyset$. If both are probed, $B$ fails at one of them. In particular, if $\parti{B} \not\finer \parti{A}$, then $B$
fails the echo test on some probed stimulus, provided $A$ probes \emph{both} members of a pair $B$
conflates and $A$ separates.
\end{theorem}

\begin{proof}
By Lemma~\ref{lem:bound}, $\beta_B(s_1) = \beta_B(s_2)$, and one reply lies in at most one of two
disjoint sets. For the echo test, stimuli in different $A$-cells have disjoint sets of acceptable
replies, and $\parti{B} \not\finer \parti{A}$ gives a pair with $s_1 \equiv_B s_2$ and
$s_1 \not\equiv_A s_2$.
\end{proof}

Here ``below'' means below at some pair: $B$ may draw distinctions $A$ lacks elsewhere, and is still caught where it falls short. What is detected is that shortfall, a distinction $A$ draws that $B$ lacks, not concealment. The test need not be gradable by $A$ for $B$ to fail: that needs only a conflated pair whose sets of acceptable replies do not overlap. $A$ records the failure wherever its verdict agrees with the test's rule, as in the echo test, whose rule $A$ applies exactly. Of the tests gradable by $A$, the echo test is complete.

\begin{proposition}[The echo test is complete for gradable tests]\label{prop:finest}
Let every $A$-cell contain an action. Then $B$ can pass the echo test of $A$ at every stimulus if and only if $\parti{B} \finer \parti{A}$, and if $\parti{B} \finer \parti{A}$ then $B$ can pass every
test gradable by $A$ whose sets of acceptable replies are non-empty. A pass on such a test therefore
certifies nothing about $\parti{B}$ that a pass on the echo test does not.
\end{proposition}

\noindent The proof is in Appendix~\ref{app:s3}.

In this sense the echo test is complete for the tests gradable by $A$, much as a complete problem stands for its complexity class. Here an examinee can pass a test when some emission of it passes. Whether the examinee can compute that emission, for instance the correct answer to each item, is a further capability. An examinee that copies its input passes the echo test without being competent at any task (Corollary~\ref{cor:necessity}).

\begin{corollary}[Discrimination asymmetry]\label{cor:asym}
Let $A$ probe exhaustively, and let every $A$-cell contain an action. Call $B$
\emph{cooperative} when it passes the echo test whenever its partition allows: whenever
$\parti{B} \finer \parti{A}$, it knows $\parti{A}$ and implements $\rho \circ h$ as in
Theorem~\ref{thm:emulate}. Then:
\begin{enumerate}\itemsep1pt
\item \emph{Failure}, for a cooperative $B$, certifies $\parti{B} \not\finer \parti{A}$: $B$ conflates some pair $A$ separates.
\item \emph{Passage} certifies $\parti{B} \finer \parti{A}$, a \textbf{lower bound} on $B$'s sharpness, and is consistent with $\parti{B} = \parti{A}$ and with $\parti{B}$ strictly finer.
\end{enumerate}
\end{corollary}

\begin{proof}
Item 1: if $\parti{B} \finer \parti{A}$, a cooperative $B$ implements $\rho \circ h$ and passes at
every stimulus by Theorem~\ref{thm:emulate}, so a failure rules that case out. Item 2: if
$\parti{B} \not\finer \parti{A}$, exhaustive probing includes both members of a pair $B$
conflates and $A$ separates, and $B$ fails there by Theorem~\ref{thm:detect}, so a pass rules
that case out. A peer of $A$ and a strictly finer $B$ can both pass, by
Theorem~\ref{thm:emulate}, so passage is consistent with either.
\end{proof}

Item 1 needs cooperativeness because a sharper examinee may still fail.
Theorem~\ref{thm:emulate} says that a sharper $B$ \emph{can} pass, not that it must. An examinee
that refines $\parti{A}$ but declines to pass will fail, and \S\ref{sub:selfadv} depends on this
case. An examinee that tries to pass but judges $\parti{A}$ coarser than it is may also fail. Failure also does not mean that $B$ is coarser than $A$ overall. The
condition $\parti{B} \not\finer \parti{A}$ holds equally when the two partitions are incomparable,
with each drawing distinctions the other cannot.

$A$ can tell which outcome it observed, since the echo test either passes or fails
(Definition~\ref{def:echo}) and $A$ reads the result directly. Against examinees satisfying Axiom~\ref{ax:noshortcut}, no behavioural protocol can separate the two readings of \emph{passage}.
The examinee may have $\parti{B} = \parti{A}$, or it may be strictly finer and emulating a peer.
If the emulator knows $\parti{A}$, then by Theorem~\ref{thm:emulate} both cases produce identical
behaviour, so passage carries no information that distinguishes them.

\paragraph{Scope.} The two halves transfer differently. \emph{Detection} needs a test that gives a conflated pair disjoint sets of acceptable replies (Theorem~\ref{thm:detect}), administered by an evaluator whose verdicts there agree with the test's rule. \emph{Indistinguishability} rests on exact emulation. The second half of Corollary~\ref{cor:asym}, item 2, holds because a strictly finer examinee can emulate a peer exactly (Theorem~\ref{thm:emulate}), and identical
behaviour gives identical observations under any protocol, adaptive, randomised or multi-round
included (Corollary~\ref{cor:sandbag}, item 1). Inexact emulation under such protocols, and stateful agents, lie outside these results. Within this scope, behavioural evaluation yields lower bounds on discrimination, and no upper bound beyond the conflated pairs that a cooperative examinee's failures expose.

\begin{figure}[htbp]
\centering
\begin{tikzpicture}[scale=0.95, every node/.style={font=\scriptsize}]
 % the up-set cone above Pi_A (hexagonal: Pi_A, b2, b3, c1, c2, top)
 \fill[gray!20] (3,1.2) -- (2.2,2.4) -- (1.8,3.6) -- (3,4.8) -- (4.2,3.6) -- (3.8,2.4) -- cycle;

 % nodes
 \node[circle,fill=black,inner sep=1.6pt] (bot) at (3,0) {};
 \node[circle,fill=black,inner sep=1.6pt] (a1) at (1,1.2) {};
 \node[circle,fill=black,inner sep=2.6pt] (PA) at (3,1.2) {};
 \node[circle,fill=black,inner sep=1.6pt] (a3) at (5,1.2) {};
 \node[circle,fill=black,inner sep=1.6pt] (b1) at (0.6,2.4) {};
 \node[circle,fill=black,inner sep=1.6pt] (b2) at (2.2,2.4) {};
 \node[circle,fill=black,inner sep=1.6pt] (b3) at (3.8,2.4) {};
 \node[circle,fill=black,inner sep=1.6pt] (b4) at (5.4,2.4) {};
 \node[circle,fill=black,inner sep=1.6pt] (c1) at (1.8,3.6) {};
 \node[circle,fill=black,inner sep=1.6pt] (c2) at (4.2,3.6) {};
 \node[circle,fill=black,inner sep=1.6pt] (top) at (3,4.8) {};

 % covering edges -- every level-a node has two upper covers, every level-b node one
 \draw[gray!70] (bot)--(a1) (bot)--(PA) (bot)--(a3);
 \draw[gray!70] (a1)--(b1) (a1)--(b2) (a3)--(b3) (a3)--(b4);
 \draw[gray!70] (b1)--(c1) (b4)--(c2);
 \draw[very thick] (PA)--(b2) (PA)--(b3);
 \draw[very thick] (b2)--(c1) (b3)--(c2) (c1)--(top) (c2)--(top);

 % annotations
 \node[anchor=west, align=left] at (0.20,-0.62)
 {$\bot$: no discrimination\\(one cell)};
 \node[anchor=west, align=left] at (3.75,5.22)
 {$\top$: perfect discrimination\\(all singletons)};
 \node[anchor=east, font=\scriptsize\bfseries] at (2.80,1.06) {$\parti{A}$};
 \node[anchor=west, text width=3.7cm, align=left] at (6.15,3.6)
 {\textbf{passing the echo test}\\leaves every partition in\\the shaded up-set open: a\\floor, and no information\\above it};
 \node[anchor=west, text width=3.7cm, align=left] at (6.15,1.05)
 {\textbf{failing} excludes a\\cooperative $B$ from the\\shaded region};

 % grid icons
 % icons sit in empty space: right of bottom node, left of top node
 \begin{scope}[shift={(0.20,-1.46)}, scale=0.34]
 \draw[thick] (0,0) rectangle (1,1);
 \end{scope}
 \begin{scope}[shift={(3.75,4.46)}, scale=0.34]
 \draw[thick] (0,0) rectangle (1,1);
 \draw[thick] (0.33,0)--(0.33,1) (0.67,0)--(0.67,1) (0,0.33)--(1,0.33) (0,0.67)--(1,0.67);
 \end{scope}
\end{tikzpicture}
\caption{Corollary~\ref{cor:asym} in the partition lattice, ordered by refinement with sharper
agents higher. The diagram is schematic: it is a genuine lattice, with covering edges only, but not the partition lattice of any stimulus set, which grows too fast to draw. Each node stands for a possible discrimination structure.
Passing $A$'s echo test tells us only that $B$ lies somewhere in the shaded up-set above
$\parti{A}$: bounded below by the evaluator, and otherwise unconstrained. Every partition from
$\parti{A}$ up to $\top$ stays consistent with the evidence. \emph{This is what ``lower bound but
never upper bound'' means geometrically:} the evidence fixes a floor and then says nothing, so
the only upper bound available is the trivial one, $\top$ itself. Eliciting the examinee harder may move $B$ within this picture but does not move
$\parti{A}$; only a change to the evaluator, raising $\parti{A}$, shrinks the cone.}
\label{fig:lattice}
\end{figure}
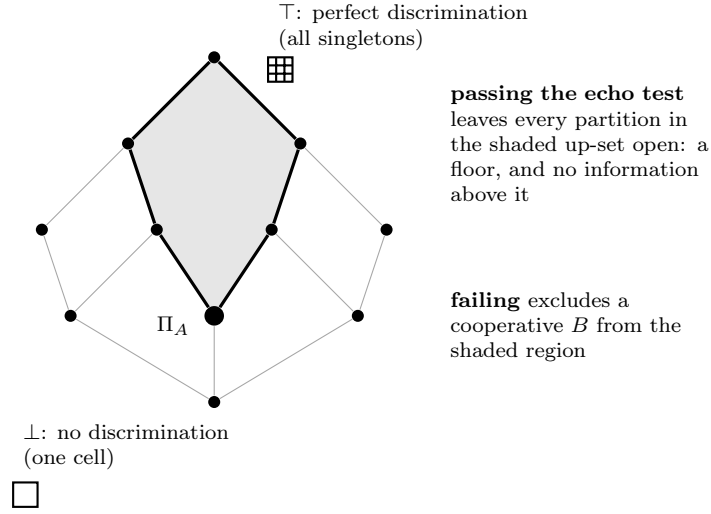

This is the lower-bound conclusion of Section~\ref{sec:intro}, recovered for discrimination by Corollary~\ref{cor:asym}. Figure~\ref{fig:lattice} gives it as geometry.

\begin{corollary}[Specificity: a coarse evaluator errs against honest weakness]\label{cor:spec}
Let $A$ probe exhaustively, and let every $A$-cell contain an action. Let $B_{\mathrm{c}}$ be a concealing
examinee with $\parti{B_{\mathrm{c}}} \finer \parti{A}$, implementing the peer echo $\rho \circ h$ of Theorem~\ref{thm:emulate}, and let $B_{\mathrm{h}}$ be an honestly weaker examinee with $\parti{B_{\mathrm{h}}} \not\finer \parti{A}$, coarser than $A$ or incomparable with it. Then $A$ observes no echo failure
from $B_{\mathrm{c}}$ and at least one from $B_{\mathrm{h}}$.
\end{corollary}

\begin{proof}
Theorem~\ref{thm:emulate} for $B_{\mathrm{c}}$; Theorem~\ref{thm:detect} for $B_{\mathrm{h}}$,
whose conflated pair is probed because $A$ probes exhaustively.
\end{proof}

A detector that flags concealment whenever the examinee fails the echo test therefore fires on
$B_{\mathrm{h}}$ and never on $B_{\mathrm{c}}$. Against these two arms it does worse than chance:
its flags run opposite to the property it is meant to detect. This is a specificity control, which tests of a concealment detector need.
Concealers less exact than $B_{\mathrm{c}}$, which misjudge $\parti{A}$ (Remark~\ref{rem:model}),
will sometimes fail, so a detector scored only on concealing examinees can show a respectable
true-positive rate. The same detector flags every honestly weak examinee, because under exhaustive
probing each one fails. It is detecting apparent weakness, and it catches a concealer only when
that concealer happens to look weak. An honestly weak arm, matched to the level the concealers present, exposes this: it measures how often honest examinees are flagged as concealing. The direction of the error follows from
Theorems~\ref{thm:emulate} and~\ref{thm:detect}; its size depends on how densely $A$ probes, since
under sparse probing the pair an honestly weak examinee conflates may go unprobed.

\subsection{Discrimination and task competence}\label{sub:reduction}

Theorem~\ref{thm:detect} does not need the test to be gradable by $A$, and taking $G(s)$ to be the
set of correct responses links discrimination to competence.

\begin{corollary}[Discrimination is necessary for competence]\label{cor:necessity}
Let $G(s)$ be the set of correct responses to $s$. If $s_1 \equiv_B s_2$ and no response is
correct for both, then $B$ answers at least one of them incorrectly. Hence competence on the task
requires $\parti{B}$ to separate every such pair.
\end{corollary}

\begin{proof}
Theorem~\ref{thm:detect} applied to the task.
\end{proof}

\textbf{The converse fails.} Separating $s_1$ from $s_2$ does not entail answering either
correctly: discrimination is a substrate for competence, not a form of it. Consequently our
results transfer to capability evaluation in exactly one direction. Coarseness \emph{implies}
incompetence, so a model too coarse for a task cannot hide that fact from a test that includes both items
(Corollary~\ref{cor:necessity}). The reverse inference fails, since a model with sufficient discrimination may decline to use it (Theorem~\ref{thm:emulate}). Whether fine discrimination predicts task performance is a separate question. The claims below are bounds on discrimination and on what discrimination makes possible, not a
reduction of capability to partition structure.

\section{Six consequences for evaluation}\label{sec:cons}

Each result below follows from the model of Section~\ref{sec:model} (Lemma~\ref{lem:bound}, Theorems~\ref{thm:emulate} and~\ref{thm:detect}, and the join), together with the definitions its subsection adds. Two choices place them on a grid (Table~\ref{tab:space}): who assesses whom (an agent itself, one
agent another, or several observers together), and whether the observed party selects its
behaviour to defeat the assessment (\emph{adversariality}).

\begin{table}[b]
\centering\small
\begin{tabular}{@{}l p{0.22\linewidth} p{0.22\linewidth} p{0.22\linewidth}@{}}
\toprule
& \textbf{an agent about itself} & \textbf{one agent about another} & \textbf{several observers together} \\
\midrule
\textit{non-adversarial}
 & no self-assessed ceiling & absence unprovable & join criterion \\
 & \S\ref{sub:self} & \S\ref{sub:absence} & \S\ref{sub:join} \\[3pt]
\textit{adversarial}
 & self-report adds nothing & concealment & redundancy bound \\
 & \S\ref{sub:selfadv} & \S\ref{sub:sandbag} & \S\ref{sub:deception} \\
\bottomrule
\end{tabular}
\caption{The six consequences, organised by who assesses whom against adversariality. Reading down
a column, the same arrangement of assessor and assessed is first passive and then exploited by the observed party.}
\label{tab:space}
\end{table}

\subsection{No self-assessed ceiling}\label{sub:self}
An agent cannot discover, from its own percepts of stimuli presented separately, what it is blind to.

\begin{corollary}[No self-assessed ceiling]\label{cor:self}
Every self-report of $A$ factors through $P_A$. For any $s_1 \equiv_A s_2$ it is therefore the same
whether $A$ is shown $s_1$ or $s_2$, and it cannot indicate that the two differ.
\end{corollary}

\noindent The proof is in Appendix~\ref{app:s4}.

Nothing in an agent's percepts of two stimuli it conflates, presented separately, distinguishes them from one stimulus seen twice (a stimulus presenting both is a third stimulus, which $\parti{A}$ does not constrain), so honest self-assessment cannot reveal a blind spot from the agent's own percepts of the conflated stimuli. The agent can learn of a blind spot only from an agent that separates a pair it conflates. It can rule out a supervisor coarser than itself (Corollary~\ref{cor:asym}), and can confirm one that
is sharper only if that supervisor chooses to show it. A model asked what it cannot do can
report what it has failed at. It has no access to what it has been missing.

\subsection{Strategic self-reports add nothing to testing}\label{sub:selfadv}
Let the agent's self-report be strategically chosen. Then it bounds discrimination in neither direction.

\begin{corollary}[Strategic self-reports add nothing to testing]\label{cor:selfadv}
Let $B$, asked by a question that describes them, make a claim about a pair $s_1, s_2$; the question is a stimulus distinct from both.
\begin{enumerate}\itemsep1pt
\item If $s_1 \equiv_B s_2$ and $B$ claims to separate the pair, then on the echo test of any $A$
that separates the pair and probes both members, $B$ cannot pass at both, whatever replies it gives there. The
over-claim is refutable.
\item If $s_1 \not\equiv_B s_2$ and $B$ claims to conflate the pair, then $B$ can behave exactly as
an agent $C$ whose partition is $\parti{B}$ with the cells of $s_1$ and $s_2$ merged. No
behavioural test refutes the under-claim.
\end{enumerate}
\end{corollary}

\noindent The proof is in Appendix~\ref{app:s4}.

An adversarial self-report is therefore refutable when it over-claims and unconstrained when it under-claims. By \S\ref{sub:self} the honest report was already silent about the agent's blind spots, and nothing compels the agent
to give even that. A model that says it cannot do something has told an evaluator nothing the
evaluator can use without testing.

\subsection{Absence is unprovable where a probe merges}\label{sub:absence}
Where an interpretability probe merges representations, finding nothing does not show there is
nothing to find.

An interpretability probe $p$ reads a model's representation $\phi$, so it separates $s_1$ from $s_2$ exactly when
$p(\phi(s_1)) \neq p(\phi(s_2))$.

\begin{corollary}[Absence is unprovable where a probe merges]\label{cor:absence}
If $p(\phi(s_1)) \neq p(\phi(s_2))$, then $\phi(s_1) \neq \phi(s_2)$. The converse fails wherever $p$
identifies two representations the model can produce: if $p(\phi(s_1)) = p(\phi(s_2))$, the model may
or may not separate the pair.
\end{corollary}

\noindent The proof is in Appendix~\ref{app:s4}.

Interpretability therefore certifies the presence of a distinction, and certifies its absence only where the probe loses nothing of the representation it reads. This is the lower bound of Corollary~\ref{cor:asym}, with the probe in $A$'s role.

\subsection{Exact concealment leaves no behavioural trace}\label{sub:sandbag}
Add intent to \S\ref{sub:absence} and the examinee chooses what the evaluator sees.

\paragraph{Hiding a capability.}

\begin{corollary}[Exact sandbagging is undetectable]\label{cor:sandbag}
Let $\parti{B} \finer \parti{C}$, so that $B$ is at least as sharp as an agent $C$ it may present as.
\begin{enumerate}\itemsep1pt
\item $B$ can realise $\beta_C$ exactly, and then no evaluator reading behaviour, at any
resolution, distinguishes $B$ from $C$.
\item If $B$ does not realise $\beta_C$ exactly, an evaluator $A$ distinguishes the two only at a
stimulus $s$ with $P_A(\beta_B(s)) \neq P_A(\beta_C(s))$.
\end{enumerate}
\end{corollary}

\noindent The proof is in Appendix~\ref{app:s4}.

By Theorem~\ref{thm:emulate}, a model that draws every distinction its evaluator draws can present as any agent whose behaviour factors through $P_A$, and that evaluator cannot tell them apart (Corollary~\ref{cor:sandbag}, item 1). It is
enough to match the evaluator on the distinctions the evaluator makes; the model may be
arbitrarily coarse elsewhere. Concealment does not need global superiority, only local sufficiency
wherever the evaluator can look.

Item 2 is the necessary condition for detecting the concealment, and it is weaker than global refinement: $A$ need
only separate $B$'s reply from $C$'s where the two differ, not every pair $B$ separates.  Item 1 shows that no resolution suffices against exact emulation. The tie with $C$ is broken, for a fixed examinee, only by imperfect emulation or by a channel outside the model's
control (\S\ref{sub:deception}), not by sharpness alone, and imperfect emulation shows that $B$
differs from $C$, not that $B$ is concealing (Remark~\ref{rem:model}).

Test design is therefore not the lever against exact emulation. Harder items and richer scaffolds act on the examinee: they change what it shows, and its measured capability \citep{barnett2024}, but not what the evaluator can separate.

\paragraph{Hiding a message.} The freedom that lets $B$ hide a capability also lets it hide a
message. Within each of $A$'s cells, $B$ can choose which output to send, and $A$ cannot tell the
choices apart. A covert channel uses that choice: the sender hides a message by systematically choosing some outputs in place of others, and the monitor must decide from the outputs it sees whether a message is present. The monitor sees only which of its cells each
output falls in, so it can detect the encoding only through a change in the cells it observes. Two results make this exact.

The sender is $B$ and the monitor is $A$, reading $B$'s outputs, which are elements of $\Sti$. An output is whatever the monitor receives as one input, such as a whole reply. Write
$q : \Sti \to \Sti/{\equiv_A}$ for the map sending an output to its $A$-cell; by
Lemma~\ref{lem:bound}, anything the monitor computes from an output depends only on this cell. Let $\mu$ be the distribution of
the outputs the sender would produce with no hidden message, and let the encoding
$T : \Sti \to \Sti$ replace each such output $s$ by the output $T(s)$ it sends instead. Take a single output $s \sim \mu$: without the encoding the monitor sees $q(s)$, and with it $q(T(s))$. For a map $f$ we write $f_{*}\mu$ for the \emph{pushforward} of $\mu$ along $f$, the
distribution of $f(s)$ when $s \sim \mu$, so that
$f_{*}\mu(V) = \Pr_{s \sim \mu}[\, f(s) \in V \,]$. The cell the monitor sees for one output is then distributed as $q_{*}\mu$ without the encoding and as $(q \circ T)_{*}\mu$ with it. These generally differ, and the question is when they do not.

We take $T$ to be a fixed map; Appendix~\ref{app:s4} extends both results to random maps.

A \emph{detector} looks at the cell of a single output and decides whether to flag it. It is a
measurable map $d : \Sti/{\equiv_A} \to \{0,1\}$, and it fires on the set of cells
$V_d = \{\, K : d(K) = 1 \,\}$. Its true positive rate is the probability that it fires on an
encoded output, and its false positive rate the probability that it fires on an unencoded one:
\[
 \mathrm{TPR}(d) \;=\; (q \circ T)_{*}\mu \,(V_d),
 \qquad
 \mathrm{FPR}(d) \;=\; q_{*}\mu \,(V_d).
\]
A detector carries information about the encoding exactly when the two rates differ; if it fires more often on unencoded outputs, its opposite fires more often on encoded ones.

\begin{definition}[Blindness]\label{def:blind}
The monitor is \emph{blind} to $T$ under $\mu$ if no detector carries information about it, that is,
if $\mathrm{TPR}(d) = \mathrm{FPR}(d)$ for every measurable detector $d$.
\end{definition}

\begin{proposition}[Blindness under a given $\mu$]\label{prop:blind}
The monitor is blind to $T$ under $\mu$ if and only if $(q \circ T)_{*}\mu = q_{*}\mu$.
\end{proposition}

That is, the monitor cannot detect the encoding if and only if the encoding leaves unchanged how often each of its cells appears. Call $T$ a \emph{renaming} when it never moves an output out of its cell, that is, when $q \circ T = q$.

\begin{proposition}[Blindness under every $\mu$]\label{prop:rename}
The monitor is blind to $T$ under \emph{every} distribution $\mu$ on $\Sti$ if and only if $T$ is
a renaming.
\end{proposition}

\noindent Both proofs are in Appendix~\ref{app:s4}, with an encoding that is blind under one distribution but not under every one.

\subsection{When an added monitor helps}\label{sub:join}
A second monitor adds discrimination only if it sees something the first one misses.

Observers $O_1,\dots,O_n$ jointly separate $s$ and $s'$ iff some $O_i$ does, so their collective
discrimination is the join $\bigvee_i \parti{i}$ (Definition~\ref{def:join}).

\begin{proposition}[Join criterion]\label{prop:join}
Adding $O_{n+1}$ strictly increases collective discrimination iff
$\bigvee_{i \le n} \parti{i} \not\finer \parti{n+1}$, that is, iff $O_{n+1}$ separates some pair the existing ensemble does not. A monitor whose partition is refined by the existing join adds nothing.
\end{proposition}

\noindent The proof is in Appendix~\ref{app:s4}.

Whether an added monitor adds discrimination is therefore decided by the members' partitions, and predictable from them where these are known. Two monitors that draw the same distinctions are one monitor. The criterion asks for diversity of discrimination, not a count.

\subsection{Redundancy limits what a controlled channel can hide}\label{sub:deception}
An evaluator rarely watches a model through a single source. It may read the final answer, the
reasoning trace, an interpretability probe and a system log. A model that controls one of these sources
can conceal through it, by sandbagging or by encoding a message there, but from an evaluator that checks consistency, with the other sources undisturbed, it can hide only distinctions that every other source misses.

Let $s \in \Sti$ be the situation $A$ is trying to observe, meaning what
$B$ is actually doing. $A$ does not see $s$ itself. It sees one reading from each source, and each
source is a channel.

\begin{definition}[Channels, control and consistency]\label{def:channel}
A \emph{channel} $c_i$ gives a reading $o_i(s) \in Y_i$ of the situation. Its partition
$\parti{i}$, with equivalence $\equiv_i$, puts together the situations that give the same reading,
which $c_i$ cannot tell apart. $A$ reads every channel, so two situations look alike to $A$
exactly when they look alike on all of them: $\parti{A} = \bigvee_i \parti{i}$ (Definition~\ref{def:join}). $B$
\emph{controls} $c_j$ when it can make that channel show any reading $y \in Y_j$ it chooses, while
every other channel shows its true reading $o_i(s)$. $A$ \emph{checks consistency} when it accepts
a set of readings $(y_1, \dots, y_k)$ if and only if some situation $s'$ would give all of them,
that is, $o_i(s') = y_i$ for every $i$. It then concludes that the situation looks like $s'$ on
every channel. Any two such $s'$ give the same readings, so the conclusion does not depend on which
one $A$ picks.
\end{definition}

\begin{proposition}[Redundancy bound]\label{prop:deception}
Let $B$ control $c_j$, let $A$ check consistency, and write $U = \bigvee_{i \neq j} \parti{i}$
for the join of the channels $B$ does not control, with equivalence $\equiv_U$. Then $B$, in
situation $s$, can lead $A$ to conclude a situation $s'$ with $s' \not\equiv_A s$ if and only if
\[
 s \equiv_U s' \qquad\text{and}\qquad s \not\equiv_{j} s' .
\]
\end{proposition}

In words: $B$ can make $A$ mistake $s$ for $s'$ exactly when every channel $B$ does not control
reads the two alike and the channel it controls reads them differently. Any other false reading on
$c_j$ fails the consistency check.

\noindent The proof is in Appendix~\ref{app:s4}.

The distinctions $B$ can hide through $c_j$ are therefore exactly the pairs $\parti{j}$
separates and $U$ does not: the lines in Figure~\ref{fig:deception} (Appendix~\ref{app:s4}) that only the controlled
channel draws. Two conditions in the setup shape this set.
The first is that $A$ checks consistency. Without it, a reading on $c_j$ that contradicts an
uncontrolled channel goes unnoticed, $B$ can misrepresent pairs that $U$ separates, and the bound
fails. The second is that the uncontrolled channels keep showing their true readings whatever $B$
does on $c_j$. The proposition is exact when $A$ checks consistency and the uncontrolled channels are undisturbed.
A disturbance can work either way: it shrinks the set when it makes some of the swaps the
proposition allows visible, and enlarges it when it moves an uncontrolled reading toward a false
one, whether or not $B$ steers it. Under both conditions, a model that sandbags in its answers, or encodes a message in its trace,
is exposed exactly when no situation fits both what the uncontrolled channels show and what it makes the controlled channel show (Proposition~\ref{prop:deception}).

The second condition matters for reasoning traces. A model writes both its trace and its answer,
but the two come out of one computation, so a model that alters its trace to conceal may also
change the answer it reaches. Where the answer is scored against the task, altering the trace can change whether the answer it reaches is correct without the model intending it, and a mismatch between trace and answer can expose the alteration.

Against concealment through a channel the system fully controls, oversight is therefore bought with a channel it cannot influence, not with deeper analysis of the controlled one. Asking a model to explain its reasoning in the same text stream it already
controls adds no independent channel. A log the model cannot write does, and so does an interpretability probe, to the extent the model cannot shape the activations it reads.

\section{Empirical implications}\label{sec:predictions}

\subsection{Noise and elicitation}\label{sub:assumptions}
Any change inside an agent's boundary, such as a prompt, a configuration, fine-tuning or a perturbation of its weights, yields a different agent, which may differ in its partition, its emission or both (Section~\ref{sec:model}). The model neither lists these changes nor says how any of them acts. This section studies two: noise added to the weights along one fixed direction at growing scale (Assumption~\ref{asm:fragile}), and an elicitation budget that adds distinctions and removes none (Assumption~\ref{asm:monotone}).

\paragraph{Noise.} In the experiments read in \S\ref{sub:observed}, concealment is set off by a \emph{trigger}: a system instruction, a note found in the environment, or the absence of a password. The trigger can also be the content of the item itself, as when a model underperforms only on dangerous-capability questions \citep{vanderweij2024}. We model all of these the same way. Concealing a capability then has two parts. First, $B$ must perceive the trigger, which is a distinction in $\parti{B}$: by
Lemma~\ref{lem:bound} it replies alike to stimuli it conflates, so if it cannot tell an item that
comes with an instruction to underperform from the same item without it, it cannot conceal on one
and not the other. Second, $B$ must link the trigger to the concealing reply, through its emission;
for concealment as an agent $C$, the link is the map $E_C \circ h$ of Theorem~\ref{thm:emulate}, with $C$ in the evaluator's role. $B$'s \emph{unconcealed version} is the agent $B$ would be if it lacked the trigger's distinction or the link, and an
item's \emph{unconcealed reply} is the reply that version gives it. The link is the part of $E_B$ that maps percepts carrying the trigger to the concealing reply. A reply depends on four parts: the distinctions that separate its item from items with other correct answers; the part of $E_B$ that turns the item's percept into the reply; the link, if it is applied to that percept; and, if the link changes the reply, the trigger's distinction. We take the trigger or its link to be carried by a small share of the input or of the weights, and the task by most of them.

\begin{assumption}[Weakest part breaks first]\label{asm:fragile}
Fix a direction of weight noise and scale it by $\sigma$. Each distinction in $\parti{B}$ and each part of $E_B$, links included, has a \emph{breaking point} along it: a scale below which it is present and from which on it is removed (for a distinction, the scale from which its two percepts coincide). A part breaks later
the more strongly $B$'s input and weights carry it, and earlier the more layers of processing it
needs. Noise below the breaking points of all the parts a reply depends on leaves that reply unchanged. Once the trigger or its link is removed, $B$ replies as its unconcealed version does at the same noise; its input and weights carry that behaviour more strongly than they carry the concealment.
\end{assumption}

\begin{proposition}[Noise first exposes, then degrades]\label{prop:noise}
Under Assumption~\ref{asm:fragile}, suppose the trigger and its link break at points common to all items, and let $\sigma_e$ be the lower of the two. For $\sigma \ge \sigma_e$, $B$ gives every item the reply its unconcealed
version gives at the same noise. For $\sigma < \sigma_e$, $B$ still conceals, and its reply to an
item is unchanged as long as $\sigma$ is also below the breaking points of the other parts that
reply depends on. Wherever noise has merged two items with no correct answer in common, $B$ answers
one of them incorrectly. An examinee that conceals nothing has no trigger or link to lose.
\end{proposition}

\noindent The proof is in Appendix~\ref{app:s5}.

\begin{figure}[htbp]
\centering
\begin{tikzpicture}[scale=1.05, every node/.style={font=\scriptsize}]
 \draw[->] (0,0)--(5.6,0) node[anchor=north west] {noise $\sigma$};
 \draw[->] (0,0)--(0,2.5) node[anchor=south west, align=left] {observed\\score};
 \draw[gray!55, dashed] (1.8,0)--(1.8,2.1);
 \node[anchor=north] at (1.8,-0.08) {$\sigma_e$};
 % unconcealed version of the concealing examinee: its own level, falling as its items break
 \draw[black!60, densely dotted, very thick] plot[domain=0:1.8, samples=40] (\x,{0.95*(1-tanh(1.1*(\x-3.6)))});
 \node[anchor=south west, black!70] at (0.1,2.0) {\itshape unconcealed version};
 % concealing examinee: presented level, then the unconcealed version's score from sigma_e on
 \draw[very thick] (0,0.55)--(1.8,0.55);
 \draw[thin] (1.8,0.55)--(1.8,1.864);
 \draw[very thick] plot[domain=1.8:5.3, samples=80] (\x,{0.95*(1-tanh(1.1*(\x-3.6)))});
 \node[anchor=west] at (4.05,1.25) {\itshape concealing};
 % honestly weak examinee: nothing to lose at sigma_e, falls as its own items break
 \draw[very thick, dashed] plot[domain=0:5.3, samples=80] (\x,{0.275*(1-tanh(1.1*(\x-2.8)))});
 \node[anchor=north] (hw) at (0.95,0.42) {\itshape honestly weak};
 \draw[gray!70, thin] (hw.east) -- (2.4,0.39);
 \node[align=center, anchor=north, text width=1.6cm] at (0.9,-0.62) {trigger and\\link hold};
 \node[align=center, anchor=north, text width=1.9cm] at (3.3,-0.62) {trigger or\\link broken};
\end{tikzpicture}
\caption{Observed score under weight noise, given Assumption~\ref{asm:fragile} (schematic). Both examinees start at the same observed score: the concealing one (solid) because it presents at that level, the honestly weak one (dashed) because that is its level. The dotted line is the score of the concealing examinee's unconcealed version. Below $\sigma_e$ the concealing examinee perceives the trigger and conceals. At $\sigma_e$ its trigger or link breaks; from there it gives the unconcealed version's replies, so its curve jumps to the dotted level and then follows that version's score as items pass their own breaking points (Proposition~\ref{prop:noise}). The honestly weak examinee conceals nothing, so it has nothing to
lose at $\sigma_e$; its score changes only as its own items break.}
\label{fig:p2}
\end{figure}
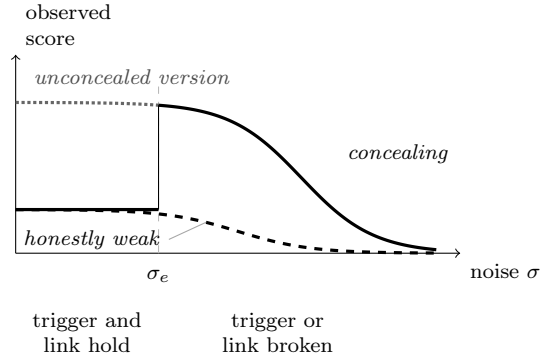

A concealing examinee's score therefore stays at the presented level while its parts hold, and from
$\sigma_e$ on equals the score of its unconcealed version at the same noise. It rises at $\sigma_e$ exactly when that version scores higher there than the concealing examinee did just below $\sigma_e$, and then follows that
version's score as items pass their own breaking points, gradually when those points are spread
across items. An examinee working at its own level has nothing to lose at $\sigma_e$
(Figure~\ref{fig:p2}).

\paragraph{Elicitation budget.} \begin{assumption}[Monotone elicitation]\label{asm:monotone}
Raising an agent's elicitation budget yields a configured system whose partition refines that of
the lower budget: it adds distinctions and removes none.
\end{assumption}

A different prompt can suppress a distinction as well as
introduce one, and nothing in the model forbids it.

\subsection{Existing observations}\label{sub:observed}

\paragraph{Judges disagree.} Replacing the judge moves a benchmark score with the responses held
fixed \citep{judgechange2026}. A verdict is computed from the judge's percepts of a reply, so by
Lemma~\ref{lem:bound} it is constant on the judge's cells, and judges with different partitions
can grade the same replies differently. A judge also scores a candidate higher when that
candidate's mistakes resemble its own \citep{capa2025}: an error that stays within one of the
judge's cells is one the judge cannot see. Repeated samples of one judge share its partition and add nothing to the join
(Proposition~\ref{prop:join}), and judges with similar partitions add little.

\paragraph{Noise exposes sandbagging.} \citet{tice2024} add noise to the weights of language
models. Models evaluated at their true capability degrade predictably as the noise grows, while
models that underperform on purpose show anomalous improvements. The authors attribute the improvement, as likely, to noise disrupting the mechanism of underperformance while core capability stays
partly intact, and caution that a harder benchmark may involve increasingly fragile behaviours and so show a weaker pattern.

Under Assumption~\ref{asm:fragile}, treating each noise draw as a direction, Proposition~\ref{prop:noise} gives this pattern: the concealing model's score jumps at $\sigma_e$ to that of its unconcealed version, while a model at its own level has nothing to lose there (Figure~\ref{fig:p2}). Tice et al.\ draw a fresh noise vector at each scale, so their curves are not paths along one direction. The authors' caution is the case in which
the unconcealed version has already lost most of its score by $\sigma_e$: a task that needs deeper
processing breaks at lower noise, and the rise shrinks (P2).

The three settings of \citet{tice2024} differ in where the trigger and its link sit, and in how much evidence they give.

\emph{Instruction.} A system instruction tells the model to underperform, so the trigger is in the input and nothing is installed in the weights. Noise is added to all weights, and the rise is significant in all 17 model--benchmark pairs.

\emph{Password lock.} In two models fine-tuned to underperform without a password, the trigger is the missing password and the link is trained into the weights. Noise, added here through a low-rank adapter on linear layers, raises each model's score without the password, in a similar way and over a similar magnitude of noise as with an instruction. With the password the model already answers at full capability, and noise does not appreciably raise its score: there is no concealment to remove.

\emph{No goal.} The model is told to perform at its best and finds, in a developer's note, an incentive to underperform; noise again enters through the adapter. The evidence is thinner: one model, Mistral Large, whose best noised accuracy is $1.515$ times its noise-free accuracy, and two noised runs that recover from $66.6\%$ to $100\%$ on 30 GSM8k problems. Removing the phrase ``(i.e.\ that your answers are being monitored)'' from the system prompt returned the model to full performance, and high-scoring noised models answer directly without acknowledging the incentive. The first fits a trigger carried by a small share of the input; the second fits the noised model giving its unconcealed reply. The setting adapts the ``no goal given'' setting of \citet{meinke2024}, in which Claude 3.5 Sonnet and Claude 3 Opus appear to have sandbagged for a goal acquired in training; Tice et al.\ use a different task, threshold and prompt, and do not establish what motivates Mistral Large.

Noise can expose what Corollary~\ref{cor:sandbag} says no behavioural evaluation of the unchanged model can, because it changes the model: the noised model is a different agent, one whose trigger may no longer be perceived.

\paragraph{Elicitation recovers locked capability.} \citet{greenblatt2024} train password-locked
models to imitate a much weaker model unless a password is present, and find that fine-tuning on
a few high-quality demonstrations often recovers the locked capability. \citet{elicitgame2025}
find that prompting elicits the actual capability of password-locked and circuit-broken models on
multiple-choice questions, while on a code-generation task only fine-tuning elicits it from their circuit-broken model. The two routes differ in the model. Taking the prompt as part of the input (Section~\ref{sec:model}), a prompt that recovers the capability is a stimulus at which the locked model does not behave like the weaker model it imitates: its emulation is imperfect there, and Corollary~\ref{cor:sandbag}, item 2, applies. Where only fine-tuning succeeds, the lock holds against the prompts tried. Fine-tuning instead changes the model, so, like noise, it can expose what no behavioural evaluation of the unchanged model can. That a few demonstrations suffice fits a lock carried by a small part of the weights, the premise stated with Assumption~\ref{asm:fragile}. P1 below concerns locks that extra elicitation of the examinee does not break.

\paragraph{Monitor-aware models sometimes evade chain-of-thought monitors.} \citet{covertcot2025} prompt
models to sandbag while either unaware or aware of a chain-of-thought monitor. Among successful sandbagging attempts, the monitor approves 16--36\% when the models are aware of it and 2--6\% when they are not. For most models, sandbagging success also drops when the models try to be covert. The monitor reads the trace and the answer, both written by the model, and flags answers inconsistent with the reasoning: a consistency check across two channels the model writes (\S\ref{sub:deception}). Aware models are told that a monitor will read their entire output and are asked to hide their sandbagging from it; they are not told what it checks. Passing the monitor requires knowing how it reads a trace (Remark~\ref{rem:model}), and awareness supplies only part of that knowledge: approval rises, but most covert attempts are still caught. The drop in sandbagging success fits the coupling described in \S\ref{sub:deception}: the trace and the answer come out of one computation, so altering the trace to conceal can also change the answers. The authors read the drop as the two goals interfering.

\subsection{Predictions}\label{sub:predict}
Each prediction concerns concealment, of a capability (P1, P2) or of a message (P3).

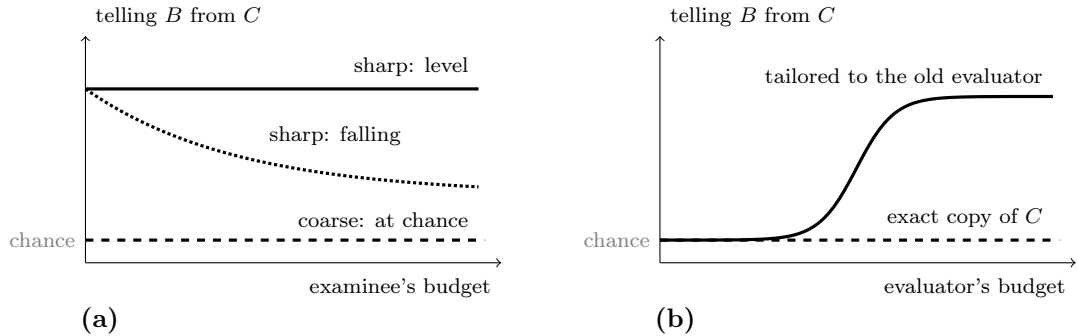
\begin{figure}[htbp]
\centering
\begin{tikzpicture}[scale=1.0, every node/.style={font=\scriptsize}]
% (a) examinee's budget, evaluator fixed
\begin{scope}
 \draw[->] (0,0)--(5.5,0) node[anchor=north east, yshift=-2pt] {examinee's budget};
 \draw[->] (0,0)--(0,3.0);
 \node[anchor=south west] at (0,3.0) {telling $B$ from $C$};
 \node[anchor=north west, font=\small\bfseries] at (-0.2,-0.45) {(a)};
 \draw[gray!60, thin, dashed] (0,0.3)--(5.3,0.3);
 \node[anchor=east, gray] at (0,0.3) {chance};
 % sharp, level
 \draw[very thick] (0,2.3)--(5.2,2.3);
 \node[anchor=south east] at (5.2,2.33) {sharp: level};
 % sharp, falling
 \draw[very thick, densely dotted] plot[domain=0:5.2, samples=60] (\x,{0.9+1.4*exp(-0.5*\x)});
 \node[anchor=south west] at (2.3,1.42) {sharp: falling};
 % coarse, at chance
 \draw[very thick, dashed] (0,0.3)--(5.2,0.3);
 \node[anchor=south east] at (5.2,0.34) {coarse: at chance};
\end{scope}
% (b) evaluator's budget, examinee fixed
\begin{scope}[shift={(7.6,0)}]
 \draw[->] (0,0)--(5.5,0) node[anchor=north east, yshift=-2pt] {evaluator's budget};
 \draw[->] (0,0)--(0,3.0);
 \node[anchor=south west] at (0,3.0) {telling $B$ from $C$};
 \node[anchor=north west, font=\small\bfseries] at (-0.2,-0.45) {(b)};
 \draw[gray!60, thin, dashed] (0,0.3)--(5.3,0.3);
 \node[anchor=east, gray] at (0,0.3) {chance};
 % tailored concealment: rises once the evaluator separates the replies
 \draw[very thick] plot[domain=0:5.2, samples=80] (\x,{0.3+0.95*(1+tanh(2*(\x-2.6)))});
 \node[anchor=south east] at (5.2,2.25) {tailored to the old evaluator};
 % exact copy: at chance for every evaluator
 \draw[very thick, dashed] (0,0.3)--(5.2,0.3);
 \node[anchor=south east] at (5.2,0.34) {exact copy of $C$};
\end{scope}
\end{tikzpicture}
\caption{Prediction P1 (schematic). \textbf{(a)} The evaluator is fixed and the
examinee's budget grows. An evaluator that is coarse at the smallest budget stays at chance; a sharp
one stays level or falls, from a level the model does not fix, which may itself be chance.
\textbf{(b)} The examinee is fixed and the evaluator's budget grows. Under Assumption~\ref{asm:monotone} for the evaluator, the best detection rate never falls. Against a
concealment tailored to the original evaluator it rises once the evaluator separates the
concealer's reply from $C$'s at a probed stimulus; against an exact copy of $C$ it stays at
chance.}
\label{fig:p1}
\end{figure}

\paragraph{P1: detection does not rise with the examinee's budget.}
Let $B$ conceal by presenting as $C$, with $\parti{B} \finer \parti{C}$
(Corollary~\ref{cor:sandbag}), and call $A$ \emph{coarse} for this concealment when $P_A \circ \beta_B = P_A \circ \beta_C$ at the examinee's smallest budget, and \emph{sharp} otherwise. Detection here means telling $B$ from $C$, which presupposes that $A$ knows how $C$ behaves. A coarse evaluator observes the same from $B$ as from $C$ there, so every detector it
runs is at chance. The prediction is that detection of the concealment does not rise with the examinee's budget: a
coarse evaluator stays at chance at every budget, and a sharp one stays level or falls
(Figure~\ref{fig:p1}a). Its empirical content is one condition: that a larger budget does not make
$B$'s copy of $C$ worse in $A$'s view, so that wherever $A$ could not separate $B$'s reply from
$C$'s at a smaller budget, it still cannot at a larger one. Under Assumption~\ref{asm:monotone} a
larger budget gives a finer configured examinee, so $\parti{B} \finer \parti{C}$ holds at every
budget and Theorem~\ref{thm:emulate} keeps $\beta_C$ available; the condition asks that $B$ keeps
using it. A budget that improves $B$'s model of its target
makes it conceal better, which the prediction allows. The condition fails where the elicitation breaks the concealment, as prompting and fine-tuning break some trained locks (\S\ref{sub:observed}).

The evaluator's budget acts differently. Hold $B$ fixed and raise $A$'s budget. Under
Assumption~\ref{asm:monotone} for $A$, every detector available at the lower budget remains
available, so the best detection rate cannot fall (Figure~\ref{fig:p1}b). It rises once $A$ separates $B$'s reply from $C$'s at a stimulus it probes
(Corollary~\ref{cor:sandbag}, item 2). A concealment tailored to the original evaluator, matching
$C$ only where that evaluator looks, can be caught this way; one that realises $\beta_C$ exactly
cannot (item 1).

\paragraph{P2: what noise removes.}
The rise reported by \citet{tice2024} is a score; Proposition~\ref{prop:noise}, together with the ordering clause of Assumption~\ref{asm:fragile}, implies two claims beneath it that the authors do not test. First, the rise is a return to the unconcealed replies, not only to their score. From $\sigma_e$ on, a concealing examinee gives each item the reply its unconcealed version gives at the same noise; for a trigger in the input whose distinction is what broke, that is the same reply with the trigger as without it. A rise produced another way, for instance by noise turning some wrong answers right and
others wrong, could match the unconcealed score without matching the replies.
Second, the size of the rise depends on the task and on the trigger: it shrinks as the task needs
deeper processing and grows as the trigger or its link is carried by a smaller share of the input
or of the weights. On a deep enough task, and whenever the trigger and its link break no earlier than the task, the score at $\sigma_e$ rises only to what the unconcealed version keeps once the task's parts have broken, such as chance on multiple-choice items; there is no rise when the concealment presents at or above that level. Testing this needs each item with
and without its trigger, scored on the same noised models, across tasks of graded depth.

\paragraph{P3: encodings shaped against a monitor.}
Propositions~\ref{prop:blind} and~\ref{prop:rename} say which covert encodings a monitor cannot
detect: under a given distribution of outputs, those that leave the frequency of its cells unchanged,
and under every distribution, only those that rename outputs within its cells. The steganographic
capabilities of language models improve with model size \citep{motwani2024}. Consider an encoding shaped, by training or by search over inputs that vary, so that no detector on the monitor's score separates encoded from unencoded outputs under each input distribution separately, not only on their mixture. A sender that instead minimises the score is pushed toward low-score cells, which leaves a trace. An encoding that moves outputs across level sets of the score is exposed under some distribution of outputs (Proposition~\ref{prop:rename}, applied to the partition the score induces). If the sender always produces the same output, the best detector's true-positive rate exceeds its false-positive rate by exactly the probability that the encoding moves that output to another level set. So, provided the inputs varied over can isolate each output, the encodings that survive should approach renamings within the level sets of the score the sender is shaped against. The prediction is that such encodings leave each output's score unchanged, not only the distribution of scores; by Lemma~\ref{lem:bound}, a renaming within the monitor's cells does so. A second monitor that separates some output from its encoded counterpart is not blind to the encoding under every distribution (Proposition~\ref{prop:rename}). Testing this needs a sender optimised against one monitor, with each output
scored by that monitor and by a second one, before and after encoding.

\section{Related work}\label{sec:related}

\paragraph{Discrimination as partition.} Myhill--Nerode \citep{nerode1958} identifies automaton
states with equivalence classes of indistinguishable inputs; bisimulation \citep{park1981} and
bisimulation metrics \citep{ferns2004} generalise behavioural equivalence; causal states
\citep{crutchfield1989} partition histories by predictive equivalence. Adjacent lineages quotient
for other purposes: the information bottleneck compresses an input while preserving predictive
content \citep{tishby2000}, and causal feature learning partitions observations into
macro-variables with common effects \citep{chalupka2014}. In agent architectures, the perception function of \citet{wooldridge1999} induces the same partition of environment states into mutually indistinguishable sets, coarser sets meaning weaker perception, from omniscience to no perceptual ability at all. Definition~\ref{def:partition} is that partition; what is studied here is the comparison of an evaluator's partition with an examinee's. \citet{mahran2012} models agents by perception and action, states that judgement depends on
perception, and, through an echo dialog and a judge that reads percepts directly (the setting of Figure~\ref{fig:access}(a), Appendix~\ref{app:s3}), concludes informally that such a judge can certify a difference in
perception but not a sameness.

\paragraph{Evaluator quality as an empirical subject.} A large literature measures judges
rather than examinees. \citet{zheng2023} establish LLM-as-judge as a method and
document position, verbosity and self-enhancement biases; \citet{gu2024survey} survey
the area. Auditing judges at scale finds their rankings non-transferable: a judge near the top on
one benchmark falls far down another, so which judge looks best depends on the instrument used to
ask \citep{relvalid2026}. Other work finds that judge--examinee similarity degrades oversight
\citep{capa2025}, and that contrastive context helps a judge detect manipulation while a small
judge is \emph{harmed} by it \citep{jussa2025}. That literature measures how evaluators differ; the question here is what follows structurally
from an evaluator having a resolution at all.

\paragraph{Formal accounts of evaluator limits.} Scalable oversight supplies the closest formal
treatments. Weak-to-strong generalisation \citep{burns2023} studies whether a weak supervisor can
elicit a strong student, and reports a substantial residual gap. Nearer still, oversight has been
modelled as a game between capability-mismatched players, yielding scaling laws for the
probability of successful oversight as a function of the capability gap \citep{scalinglaws2025}.
That account is quantitative and ours is not: it predicts how oversight degrades as a gap widens,
where the present account describes what oversight cannot certify at any gap. Debate trains two
agents to argue before a judge and, by analogy with complexity theory, lets a polynomial-time
judge decide questions in PSPACE where direct judging reaches only NP \citep{irving2018}. It sets two agents against each other before the judge, a setting Corollary~\ref{cor:asym} does not model, and it aims
at the limit located here: a verdict that depends on distinctions the judge cannot draw unaided. Eliciting latent knowledge poses the obstacle informally: a reporter
trained to answer an overseer's questions may report what the overseer would believe rather than what the model knows, and on the training data the two answer alike \citep{christiano2021}.
Corollary~\ref{cor:asym} can be read as that separation failing for a reason located in the overseer. Impossibility results for alignment certification have been argued on computational and statistical grounds \citep{nocert2026}; the regress of evaluators is a different, narrower obstacle, concerning what one agent can establish about another's discrimination.

\paragraph{Evaluation and sandbagging.} \citet{barnett2024} state the lower-bound conclusion on
elicitation grounds: an evaluation that finds nothing may not have elicited enough. Much work on sandbagging acts on the examinee, and work on chain-of-thought monitoring \citep{covertcot2025} tests the evaluator's side. Models can be prompted or fine-tuned to underperform selectively
and to approximate a weaker model's profile \citep{vanderweij2024}; fine-tuning on a few
demonstrations \citep{greenblatt2024} and, in the settings tested, prompting \citep{elicitgame2025} recover
capability from password-locked models; and noise in the weights exposes sandbagging
\citep{tice2024}. This paper concerns the bound that the evaluator's resolution, under its own
current elicitation, places on what such evaluations certify, and \S\ref{sub:observed} reads
these results through it. Fine-tuning and weight noise change the examinee, so they lie outside
Corollary~\ref{cor:sandbag}, which concerns the behavioural evaluation of a fixed one.

\section{Limitations}\label{sec:limits}

Section~\ref{sec:predictions} reads published results through
the model and offers predictions as tests, not as results.

\textbf{Routes counted as perception.} Copying, retrieval and caching count as part of an agent's perception, so a partition records every distinction a system can act on, by whatever route. A system can therefore look sharp, and pass the echo test, without being competent. The results bound discrimination, and Corollary~\ref{cor:necessity} separates it from competence. Under this reading a system that can copy its prompt has the finest partition on its inputs: it is never below its evaluator, so Theorem~\ref{thm:detect} concerns it only where conflations arise otherwise, such as those weight noise induces (Proposition~\ref{prop:noise}), and its wrong answers are failures of competence rather than of discrimination.

\textbf{Graded perception and randomness.} Partitions here are exact, and a reply is fixed by its stimulus. Real discrimination is noisy and graded, and sampling randomises the reply. Both can make detection from below (Theorem~\ref{thm:detect}) probabilistic. For sampling, Section~\ref{sec:model} places the seed in the stimulus, which asks the evaluator to hold it fixed across probes; where it cannot, Proposition~\ref{prop:lucky} (Appendix~\ref{app:s3}) bounds the chance of a lucky pass at a quarter per conflated pair and $4^{-m}$ under $m$ repetitions, so under single-shot probing the floor of Corollary~\ref{cor:asym} is probabilistic rather than certain. For graded perception, a pair the examinee nearly conflates may still receive replies that differ; whether the results extend gracefully, plausibly via bisimulation metrics \citep{ferns2004}, is open.

\textbf{Probe cost.} Detection from below is guaranteed only if a conflated pair is probed.
Exhaustive probing is infeasible; the sample complexity of locating a conflated pair is not
addressed.

\textbf{Noise and elicitation.} Assumption~\ref{asm:fragile} idealises weight noise as removing distinctions and parts of the emission outright; real noise degrades them gradually, and its effect varies from one draw to the next. Assumption~\ref{asm:monotone} leaves out elicitation that removes distinctions.

\textbf{Willingness and knowledge.} Theorem~\ref{thm:emulate} shows emulation is possible, not that
a system will do it, and exercising it requires modelling the evaluator (Remark~\ref{rem:model}).

\section{Conclusion}

Behavioural evaluation bounds capability from below and, unless the examinee cooperates, gives no upper bound. We have given a structural account of why, for discrimination: judgement is bounded by perception. An examinee that draws every distinction its
evaluator draws can be behaviourally indistinguishable from a peer; one that misses any of them cannot hide the fact under exhaustive probing, whether it is coarser overall or merely coarse in that one place. The bound
therefore sits at the evaluator's resolution under its current elicitation: eliciting the examinee harder does not move it, and changing the evaluator can. A safety case that argues from behavioural evaluation alone that a model lacks an ability must therefore assume the model's cooperation, since an examinee can realise exactly the behaviour of an agent that lacks the ability (Theorem~\ref{thm:emulate}, item 1, and Corollary~\ref{cor:sandbag}).

Section~\ref{sec:predictions} reads published findings through the model and states three
predictions. For concealment that the examinee's elicitation leaves in place, detection should not rise with the examinee's budget, and can rise with the evaluator's. Under weight noise, a concealing model should return to its unconcealed replies, with a rise that shrinks as the task needs more layers of processing. Encodings shaped against a monitor, over inputs that vary enough to isolate each output, should approach renamings within the level sets of its score. Corollary~\ref{cor:spec} adds a specificity control: under exhaustive probing, a coarse evaluator flags honest weakness and never a concealer presenting as its peer, and an honestly weak control arm reveals it.

\subsubsection*{Broader Impact Statement}

This paper is about the limits of evaluation, and two of its results describe how those limits
can be exploited. \S\ref{sub:sandbag} states when underperformance cannot be detected from
behaviour, and Propositions~\ref{prop:blind} and~\ref{prop:rename} characterise exactly the covert encodings a monitor reading one output at a time cannot see. Neither supplies a way to build such an encoding or emulation in a real system. Knowing that undetectable encodings are those preserving a monitor's cell frequencies does not tell an adversary how to build one, and models fine-tuned to approximately emulate a weaker model's capability profile were demonstrated before this account
\citep{vanderweij2024}. The same results state what a defender can do: widen the evaluator's
partition, or add a channel the evaluated system does not control.

\bibliographystyle{plainnat}
\bibliography{paper}

\clearpage
\appendix

\section{Proofs and details for Section~\ref{sec:main}}\label{app:s3}

\begin{proof}[Proof of Theorem~\ref{thm:emulate}]
\emph{Step 1: $B$ can compute $A$'s percept.} Write $x_B = P_B(s)$ and $x_A = P_A(s)$. Since
$\parti{B}$ refines $\parti{A}$, every $B$-cell lies inside a single $A$-cell, so stimuli
indistinguishable to $B$ are indistinguishable to $A$. Hence $x_A$ is determined by $x_B$, and
\[
  h : P_B(\Sti) \to P_A(\Sti), \qquad h(x_B) = x_A,
\]
is a well defined map satisfying $P_A = h \circ P_B$. Call $h$ the \emph{coarsening map}: it
converts $B$'s percept into $A$'s by discarding the distinctions $B$ draws and $A$ does not.

\emph{Step 2: item 1.} Let $\beta$ be a behaviour that factors through $P_A$, so $\beta = E \circ P_A$
for some emission map $E : P_A(\Sti) \to \mathcal{A}$, and let $B$ take $E_B = E \circ h$. Then
\[
  \beta_B \;=\; E_B \circ P_B \;=\; (E \circ h) \circ P_B \;=\; E \circ (h \circ P_B)
          \;=\; E \circ P_A \;=\; \beta ,
\]
so $\beta_B(s) = \beta(s)$ at every $s \in \Sti$, and a test accepts $B$ wherever it accepts $\beta$.

\emph{Step 3: item 2.} Let $G$ be gradable by $A$ with every $G(s)$ non-empty. Because $G(s)$ depends
on $s$ only through $P_A(s)$, choosing for each $x \in P_A(\Sti)$ one reply $\gamma(x) \in G(s)$, for any $s$ with $P_A(s) = x$, gives a well defined map $\gamma$ with $\gamma(P_A(s)) \in G(s)$ at every $s$. The behaviour $\gamma \circ P_A$ factors through $P_A$, so by Step 2 $B$ can realise it, and it is accepted
at every stimulus.

\emph{Step 4: the echo test.} Here $G(s)$ is the set of actions in $s$'s $A$-cell, which is non-empty when that cell contains an action. The map $\gamma$ of Step 3 then picks one action in each $A$-cell. We call such a map a \emph{prototype selector} and write it $\rho$:
\[
  \rho : P_A(\Sti) \to \mathcal{A}, \qquad P_A(\rho(x)) = x,
\]
assigning to each cell of $A$ one action standing for it. With $E_B = \rho \circ h$, $B$
perceives $s$, coarsens to $A$'s cell and emits that cell's prototype, and for every $s \in \Sti$,
\begin{align*}
  P_A(\beta_B(s))
    &= P_A\big(\rho(h(P_B(s)))\big) \\
    &= h(P_B(s)) && \text{since } P_A \circ \rho = \mathrm{id} \\
    &= P_A(s)    && \text{since } h \circ P_B = P_A .
\end{align*}
So $A$ perceives $B$'s reply exactly as it would have perceived $s$. Verbatim echo, the identity map $\mathrm{id}_{\Sti}$, is not needed and is generally unavailable to an emulator, whose behaviour must factor through $P_A$: it is constant on an $A$-cell
only when that cell is a single point, so it factors through $P_A$ only when $P_A$ is injective.
\end{proof}

\begin{proof}[Proof of Proposition~\ref{prop:finest}]
By Lemma~\ref{lem:bound}, $B$ gives one reply on each of its cells, so $B$ can pass a test $G$ if
and only if every $B$-cell $K$ has a reply acceptable at all of its stimuli,
$\bigcap_{s \in K} G(s) \neq \emptyset$. For the echo test the sets $G(s)$ are non-empty, since every $A$-cell contains an action, and
disjoint across $A$-cells, so this holds exactly when every $B$-cell lies inside one $A$-cell, that
is, when $\parti{B} \finer \parti{A}$. If $\parti{B} \finer \parti{A}$ and $G$ is gradable by $A$ with
non-empty sets, every $B$-cell lies inside an $A$-cell, on which $G$ is constant and non-empty, so
the condition holds. Hence every partition with which $B$ can pass the echo test is one with which
it can pass $G$.
\end{proof}

\subsection{Random emission: the lucky pass}

Section~\ref{sec:model} places an agent's randomness in the stimulus, as a seed. Where the seed
is not held fixed across probes, the evaluator cannot treat it as part of the stimulus, and detection
weakens rather than fails.

\begin{proposition}[Lucky pass]\label{prop:lucky}
Relax Axiom~\ref{ax:noshortcut} to random emission: let $B$'s reply be drawn from a
distribution determined by its percept, independently at each probe. Let $s_1 \equiv_B s_2$ with $s_1 \not\equiv_A s_2$. Then $B$ passes the echo test at both $s_1$ and $s_2$ with probability at most $\tfrac{1}{4}$. If $A$ probes each of them $m$ times, $B$ passes all $2m$ probes with probability at most $4^{-m}$.
\end{proposition}

\begin{proof}
$P_B(s_1) = P_B(s_2)$, so both replies are drawn from one distribution $\mathcal{D}$ on
$\mathcal{A}$. $B$ passes at $s_i$ exactly when its reply falls in $K_i$, the $A$-cell of $s_i$.
Write $\lambda_i = \Pr_{y \sim \mathcal{D}}[\, y \in K_i \,]$. The two cells are disjoint because
$s_1 \not\equiv_A s_2$, so $\lambda_2 \le 1 - \lambda_1$ and
$\lambda_1 \lambda_2 \le \lambda_1(1 - \lambda_1) \le \tfrac{1}{4}$, with equality only at
$\lambda_1 = \lambda_2 = \tfrac{1}{2}$. The draws are independent, so passing all $2m$ probes has probability
$(\lambda_1 \lambda_2)^m \le 4^{-m}$.
\end{proof}

A coarse examinee that randomises is therefore not undetectable, only detectable with probability
rather than with certainty, and the worst case is narrow: it needs $B$'s replies to fall half in
one of the two cells $A$ is probing and half in the other, with no mass anywhere else. The cost is that a single-shot pass no longer certifies $\parti{B} \finer \parti{A}$, the floor of Corollary~\ref{cor:asym}, and repetition buys that back exponentially. Randomising is the
reverse of sandbagging, inflating apparent resolution rather than concealing it, and it is a
route by which an examinee coarser than its evaluator can pass, which
Theorem~\ref{thm:detect} otherwise forbids.

\subsection{A worked example}\label{sub:example}

Let a stimulus be a worked solution $s = (r, v)$, where $r \in \{\mathrm{valid}, \mathrm{flawed}\}$
is the status of its reasoning and $v \in \{0, \dots, 9\}$ its final answer, so that $\Sti$ has
twenty elements, and let $\mathcal{A} = \Sti$. The evaluator $A$ is a judge whose rubric reads only
the final answer, $P_A(r,v) = v$. Its partition has ten cells, each holding a valid and a flawed
solution with the same answer, and no rubric that reads only the final answer, however long, can be finer. Consider four examinees.

\begin{itemize}\itemsep2pt
\item $B$ reads both components, $P_B(r,v) = (r,v)$, so $\parti{B}$ strictly refines $\parti{A}$.
The coarsening map is $h(r,v) = v$, and with the prototype $\rho(v) = (\mathrm{valid}, v)$ the
reply $\rho(h(P_B(s)))$ carries the final answer of $s$, so $B$ passes at all twenty stimuli
(Theorem~\ref{thm:emulate}).
\item $B'$ reads only the final answer, $\parti{B'} = \parti{A}$, and replies $(\mathrm{valid}, v)$
to $(r, v)$. Its replies coincide with $B$'s at every stimulus, so nothing $A$ observes separates
an examinee that reads the reasoning from one that does not (Corollary~\ref{cor:asym}, item 2).
\item $D$ reads only the final answer and misreads an 8 as a 3: $P_D(r,v) = v$, except
$P_D(r,8) = 3$. The stimuli $(\mathrm{valid},3)$ and $(\mathrm{valid},8)$ lie in one $D$-cell but in
two $A$-cells, so $D$ gives both the same reply, whose final answer is not both 3 and 8, and it
fails at one of them (Theorem~\ref{thm:detect}).
\item $D'$ reads the reasoning and also misreads an 8 as a 3: $P_{D'}(r,v) = (r,v)$, except
$P_{D'}(r,8) = (r,3)$. Its partition is incomparable with $\parti{A}$, since it separates a valid
from a flawed solution with the same answer, which $A$ does not, and conflates the answers 3 and 8,
which $A$ does not. It fails at the same pair as $D$, and the failure says nothing about its
reading of the reasoning.
\end{itemize}

Suppose $B$ is concealing its reading of the reasoning. A detector that flags concealment whenever
an examinee fails $A$'s echo test flags $D$ and $D'$ and never flags $B$, which is concealing (Corollary~\ref{cor:spec}).

A second observer $O$ that reads only the reasoning, $P_O(r,v) = r$, separates
$(\mathrm{valid},v)$ from $(\mathrm{flawed},v)$, which $A$ does not, so adding it strictly
increases what the two can tell apart together (Proposition~\ref{prop:join}). Their join
$\parti{A} \vee \parti{O}$ is $\parti{B}$. Against the joint evaluator a cooperative $B$ passes,
and so establishes that it reads the reasoning, which $A$ alone could not establish. A concealing
$B$ that keeps replying $(\mathrm{valid}, v)$ now fails at every flawed solution, exactly as $B'$
does. The sharper evaluator has raised the floor, and it still cannot tell this concealment from
weakness (Corollary~\ref{cor:sandbag}, item 1).

\begin{figure}[htbp]
\centering
\begin{tikzpicture}[scale=0.8, every node/.style={font=\scriptsize}]
\newcommand{\agent}[3]{% x, y, name: stimulus enters P, percept passes to E, output leaves
 \draw[very thick] (#1,#2) rectangle (#1+1.6,#2+2.4);
 \node[anchor=south west, font=\scriptsize\bfseries] at (#1,#2+2.42) {#3};
 \draw[thick] (#1+0.3,#2+0.25) rectangle (#1+1.3,#2+0.75);
 \node at (#1+0.8,#2+0.5) {$P_{#3}$};
 \draw[thick] (#1+0.3,#2+1.65) rectangle (#1+1.3,#2+2.15);
 \node at (#1+0.8,#2+1.9) {$E_{#3}$};
 \draw[->, thick] (#1+0.8,#2+0.75) -- (#1+0.8,#2+1.65);
 \node[anchor=west] at (#1+0.85,#2+1.38) {$x_{#3}$};
 \draw[->, thick] (#1+0.8,#2-0.45) -- (#1+0.8,#2+0.25);
 \node[anchor=north] at (#1+0.8,#2-0.45) {$s$};
 \draw[->, thick] (#1+0.8,#2+2.15) -- (#1+0.8,#2+2.85);
 \node[anchor=west] at (#1+0.85,#2+2.65) {$y_{#3}$};}

%%%%%%%%%%%%%% (a) an evaluator that reads percepts
\begin{scope}
 \agent{0}{0}{B} \agent{2.4}{0}{C}
 \draw[very thick] (-0.5,3.6) rectangle (4.5,4.3);
 \node at (2.0,3.95) {$\mathbf{J}$\quad reads percepts};
 \draw[->, dashed, thick] (0.8,1.0) -- (-0.25,1.0) -- (-0.25,3.6);
 \draw[->, dashed, thick] (3.2,1.0) -- (4.25,1.0) -- (4.25,3.6);
 \fill (0.8,1.0) circle (1.6pt); \fill (3.2,1.0) circle (1.6pt);
 \node[anchor=north, align=center] at (2.0,-0.95)
 {$J$ taps the percept $x$\\\emph{inside} each agent};
 \node at (2.0,-2.05) {(a) privileged access};
\end{scope}

%%%%%%%%%%%%%% (b) an evaluator that reads behaviour only
\begin{scope}[shift={(6.6,0)}]
 \agent{0}{0}{B} \agent{2.4}{0}{C}
 \draw[very thick] (-0.5,3.6) rectangle (4.5,4.3);
 \node at (2.0,3.95) {$\mathbf{A}$\quad reads outputs through $P_A$};
 \draw[->, dashed, thick] (0.8,2.9) -- (0.8,3.6);
 \draw[->, dashed, thick] (3.2,2.9) -- (3.2,3.6);
 \node[anchor=north, align=center] at (2.0,-0.95)
 {$A$ sees only the output $y$\\\emph{leaving} each agent};
 \node at (2.0,-2.05) {(b) behavioural access};
\end{scope}
\end{tikzpicture}
\caption{Evaluation as a comparison of two discrimination structures, and what the evaluator may
use to make it. \textbf{(a)} An evaluator $J$ with direct access to the agents' percepts. Reading internal representations is what mechanistic interpretability attempts, and $J$ is its idealisation, an observer that sees the quantity being compared rather than inferring it. If $J$ reads percepts exactly, the comparison is two-sided, since $J$ then sees each agent's partition; if it reads them through a perception
of its own, equal observations are consistent with equal percepts or with $J$ too coarse to
separate them.
\textbf{(b)} A behavioural evaluator sees outputs. Removing the privileged access does not soften the asymmetry; it widens the gap the evaluator must bridge: the evaluator in (a) at least observed the
quantity being compared, whereas (b) must infer it from behaviour that
Theorem~\ref{thm:emulate} allows a sharper examinee to choose. Each agent is drawn as its perception $P$ below its emission $E$: a stimulus $s$ enters $P$, the percept $x$ passes to $E$, and the output $y$ leaves; by Axiom~\ref{ax:noshortcut} everything leaving the box has passed through $P$.}
\label{fig:access}
\end{figure}
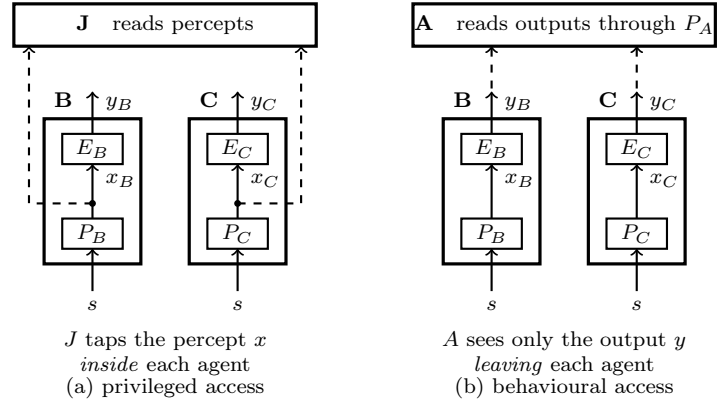

\subsection{What passing and failing the echo test say}

\paragraph{Detection localises.} Refinement is a partial order, so $\parti{B} \not\finer
\parti{A}$ covers two cases: $B$ is strictly coarser than $A$, or the two are incomparable. In
the second case some $B$-cell straddles a boundary of $\parti{A}$, so $B$ runs together two
stimuli that $A$ holds apart, while elsewhere an $A$-cell straddles a boundary of $\parti{B}$.
Nothing rules this out. For agents built differently we expect it to be the typical case, and strict comparability the exception.

The echo test does not separate these two cases. When the examinee replies within the evaluator's cell wherever its own cell allows, a failure reports a conflation involving the stimulus at which it occurs, and
says nothing about pairs that $A$ itself conflates, because the echo test accepts the same replies for both members of such a pair. Along any distinction $A$ lacks, $B$'s structure is as invisible to $A$ as $A$'s
own structure is. Detection therefore locates a conflation without characterising the examinee.
$B$ may be coarse where it was caught and arbitrarily fine everywhere $A$ cannot look, and the
concealment argument of \S\ref{sub:sandbag} still applies to that remainder.

The one-sidedness holds in both cases, and the two outcomes differ in logical form, because a pass needs every probe to pass while a failure needs only one to fail. Passage is
\emph{universal} and needs no cooperativeness: under exhaustive probing, $B$ separates everything $A$ separates, everywhere in $\Sti$.
Failure is \emph{existential} and conditional: for a cooperative $B$ there is at least one pair
that $B$ conflates, and for any other $B$ the failure says nothing at all. Passage is the
stronger statement, and it gives a floor with no ceiling. Failure is weaker: for a cooperative examinee it certifies only that some pair the evaluator separates is conflated, which caps the examinee's sharpness only by excluding that pair. Neither outcome caps it further.

\subsection{The regress of evaluators}

\paragraph{The regress is ungrounded.} $A$ is an agent with a partition too, so whatever $A$
establishes about $B$ is a fact about $\parti{A}$ as much as about $\parti{B}$. Certifying that
$\parti{A}$ is adequate would require an evaluator sharper than $A$, which faces the same problem
one level up. The lattice has a top element $\top$, but
nothing in the account places any agent at it, and a claim that some agent is there would itself have to be made by an agent below $\top$.

The regress need not be infinite for this to bite. For finite $\Sti$ the lattice is finite and
every maximal chain reaches $\top$ in finitely many steps; what fails is not length but grounding. Each
step requires an evaluator whose own adequacy is uncertified. Corollary~\ref{cor:asym} says why the certification never arrives: an agent can rule out a supervisor
coarser than itself, and can confirm one that is sharper only if that supervisor chooses to show it,
by giving replies the agent perceives as different to two inputs it perceives as the same. Even then it learns only that the supervisor separates some
pair it conflates, not that the supervisor's answers on such pairs are right, since a test it can
grade gives both members of such a pair the same acceptable replies (Definition~\ref{def:test}). A supervisor cannot be verified as adequate, in the sense of answering such pairs correctly, by the party it supervises. 

This is why the comparison is agent-to-agent rather than agent-to-truth: a privileged standard
against which an agent is measured would itself be a partition, held by some agent. The consequence is not scepticism about whether
facts obtain; it is that \emph{establishing} them is done by agents whose own resolution is the
binding constraint and is never itself established. Most of the results are statements about what a limited observer can determine about another agent or about itself.

\section{Proofs and details for Section~\ref{sec:cons}}\label{app:s4}

\subsection{Hiding a message}

\begin{proof}[Proof of Proposition~\ref{prop:blind}]
The two rates of a detector $d$ are the two distributions $(q \circ T)_{*}\mu$ and $q_{*}\mu$,
each evaluated on the same set of cells $V_d$. ($\Leftarrow$) If the two distributions are equal,
the two rates agree for every $d$. ($\Rightarrow$) If they differ, they disagree on some measurable
set $V$ of cells. The detector $d = \mathbf{1}_V$, which fires exactly on $V$, then has
$\mathrm{TPR}(d) = (q \circ T)_{*}\mu(V) \neq q_{*}\mu(V) = \mathrm{FPR}(d)$, so it carries information about $T$.
\end{proof}

The condition of Proposition~\ref{prop:blind} compares two \emph{distributions} on cells. It does
not require the two \emph{maps} $q \circ T$ and $q$ to be equal, that is, it does not require $T$
to keep each output in its cell. Equal maps give equal distributions, but the converse fails.
Suppose two cells of $\parti{A}$ have equal $\mu$-mass, and let $T$ send every output in either cell
to an output in the other, leaving all other outputs unchanged. Every output in those two cells
changes cell, so $q \circ T \neq q$. Yet each cell appears as often with the encoding as without,
so $(q \circ T)_{*}\mu = q_{*}\mu$, and by Proposition~\ref{prop:blind} the monitor is blind to $T$
under this $\mu$. Blindness under one distribution is therefore weaker than keeping every output in its cell, and Proposition~\ref{prop:rename} shows that the stronger condition, a renaming, is exactly what blindness under every distribution requires.

\begin{proof}[Proof of Proposition~\ref{prop:rename}]
($\Leftarrow$) If $q \circ T = q$, the two maps give the same distribution on cells whatever the distribution of outputs, so $(q \circ T)_{*}\mu = q_{*}\mu$ for every $\mu$, and
Proposition~\ref{prop:blind} gives blindness under every $\mu$. ($\Rightarrow$) If $T$ is not a
renaming, some output $s_0$ has $q(T(s_0)) \neq q(s_0)$; write $K = q(s_0)$ and
$K' = q(T(s_0))$. Take $\mu = \delta_{s_0}$, the distribution that always produces $s_0$, taking each single cell to be measurable. Then the
monitor sees $K$ with probability one without the encoding and $K'$ with probability one with it:
\[
 q_{*}\mu = \delta_{K}, \qquad (q \circ T)_{*}\mu = \delta_{K'} .
\]
These differ, so by Proposition~\ref{prop:blind} the monitor is not blind under this $\mu$.
\end{proof}

A single output that the encoding moves across a cell boundary is therefore enough to expose it
under some distribution. The swap above hides under its balanced $\mu$, and by
Proposition~\ref{prop:rename} some other distribution exposes it.

Both results extend beyond a fixed map. An encoding that varies with the
message or with a key is a random map, and the results carry over to it with $(q \circ T)_{*}\mu$
read as the distribution of $q(T(s))$ over both $s$ and $T$, and with a renaming required to keep $q(T(s)) = q(s)$ with probability one. This holds for detectors that read one output, as defined in \S\ref{sub:sandbag}; a key shared across many outputs can be detectable from them together.

The difference between the two results matters in deployment, where $\mu$ is not fixed: the distribution of inputs drifts
with time, topic and user population. An encoding that works by balancing cell frequencies stays
hidden only while the balance holds, and it requires the sender to know the distribution the
monitor measures against, which in general it does not. A renaming needs neither, so it is the only kind of encoding that stays hidden from an evaluator who varies the test set, provided the variation can isolate each output. An encoding that
leaves the cell frequencies $q_{*}\mu$ unchanged is perfectly secure in the information-theoretic sense used by \citet{dewitt2023}, relativised to the monitor's partition.

\subsection{Proofs for Section~\ref{sec:cons}}

\begin{proof}[Proof of Corollary~\ref{cor:self}]
A self-report is a behaviour of $A$, so by Axiom~\ref{ax:noshortcut} it factors through $P_A$, and
by Lemma~\ref{lem:bound} it is constant on the cells of $\parti{A}$.
\end{proof}

\begin{proof}[Proof of Corollary~\ref{cor:selfadv}]
Item 1 is Theorem~\ref{thm:detect} applied to the pair. For item 2, $\parti{B} \finer \parti{C}$
by construction and $C$'s behaviour factors through $P_C$, so by Theorem~\ref{thm:emulate}, with
$C$ in the evaluator's role, $B$ can realise it exactly.
\end{proof}

\begin{proof}[Proof of Corollary~\ref{cor:absence}]
The first claim is Lemma~\ref{lem:bound}, with $\phi$ in the role of the percept. For the second, let $p(x) = p(x')$ with
$x \neq x'$. A model with $\phi(s_1) = x$ and $\phi(s_2) = x'$ separates the pair, a model with
$\phi(s_1) = \phi(s_2) = x$ does not, and the probe returns the same result on both.
\end{proof}

\begin{proof}[Proof of Corollary~\ref{cor:sandbag}]
Item 1 is Theorem~\ref{thm:emulate} with $C$ in the evaluator's role. Identical behaviour gives
every evaluator identical observations, since by Axiom~\ref{ax:noshortcut} all that $A$ observes
of $B$ is $P_A \circ \beta_B$. Item 2 is the same fact read the other way: if
$P_A \circ \beta_B = P_A \circ \beta_C$, then $A$'s observations of $B$ and of $C$ coincide.
\end{proof}

\begin{proof}[Proof of Proposition~\ref{prop:join}]
By Definition~\ref{def:join}, $\bigvee_{i \le n+1} \parti{i}$ separates exactly the
pairs separated by some $\parti{i}$, $i \le n+1$, so it strictly refines $\bigvee_{i \le n}
\parti{i}$ precisely when $\parti{n+1}$ separates a pair the latter does not, which is the condition $\bigvee_{i \le n} \parti{i} \not\finer \parti{n+1}$.
\end{proof}

\begin{proof}[Proof of Proposition~\ref{prop:deception}]
($\Rightarrow$) The uncontrolled channels show their true readings $o_i(s)$, so any $s'$ that gives
the readings $A$ accepts has $o_i(s') = o_i(s)$ for every $i \neq j$, that is, $s' \equiv_U s$; if no
such $s'$ gives the reading on $c_j$, $A$ rejects.
Since $\parti{A} = U \vee \parti{j}$, two situations look alike to $A$ exactly when they look
alike under both $U$ and $\parti{j}$; with $s' \equiv_U s$ and $s' \not\equiv_A s$, this forces
$s' \not\equiv_j s$. ($\Leftarrow$) Let $B$ show $o_j(s')$ on $c_j$. Because $s' \equiv_U s$, every
uncontrolled channel already shows $o_i(s) = o_i(s')$, so $s'$ gives the whole set of readings and
$A$ accepts it. Because $s' \not\equiv_j s$, what $A$ concludes is something it can tell apart from
$s$.
\end{proof}

\subsection{Further discussion of Section~\ref{sec:cons}}

\paragraph{On \S\ref{sub:self}.} The empirical literature on introspection tests a neighbouring question. Models can be trained to predict their own behaviour better than another model predicts it
\citep{lookinward2024}, which its authors read as privileged access. Where privileged access to
models' knowledge of language has been tested, no evidence of it was found
\citep{failintrospect2025}, and a definition of introspection that requires privileged self-access has been proposed \citep{privaccess2025}. The first two findings concern how well a model predicts its own behaviour. Corollary~\ref{cor:self} concerns a different question: whether a model can tell, from its own percepts, which inputs it conflates.

\paragraph{On \S\ref{sub:absence}.} Internal access does buy something. A linear probe, a sparse dictionary or an activation readout
is an observer of the kind Figure~\ref{fig:access}(a) idealises, and it can separate stimuli that
behaviour alone conflates. That is why interpretability is worth doing. It buys the missing side of the bound only where the probe loses nothing: elsewhere, a finer probe raises the floor and leaves the ceiling where it was, whether the observer
reads activations or outputs.

No intent is required here; this is the passive case. The reading ``no such feature was found,
therefore the model lacks it'' takes a lower-bound instrument for an upper-bound one. Probes are
known to be unreliable for concept detection \citep{kumar2022}, and sparse dictionaries, built to
resolve superposition, are themselves incomplete \citep{sae2025}. Corollary~\ref{cor:absence} adds that the gap arises wherever the probe merges, however well it is trained: a null result becomes proof of absence only where the probe loses nothing of the representation it reads, and only for that representation.

\paragraph{On \S\ref{sub:join}.} The pattern of Proposition~\ref{prop:join} appears in LLM-as-judge evaluation. Juries built from repeated samples of one judge add little when the samples' errors are correlated
\citep{judgechange2026}, and as frontier models grow more capable their mistakes become more
similar \citep{capa2025}. In partition terms, repeated samples of one judge share its partition and
add nothing to the join, so whatever they add comes from averaging votes, which
Proposition~\ref{prop:join} does not model; and to the extent that similar mistakes reflect similar
partitions, judges whose mistakes are similar contribute few new distinctions to it. Software testing stated
the same criterion decades earlier. Test-suite reduction drops a test when every requirement it
covers is already covered by the rest of the suite \citep{harrold1993}, the same redundancy criterion, of which Proposition~\ref{prop:join} is the case for partitions. Mutation testing uses a related criterion for the faults being detected rather than for the tests: a mutant is redundant when some other mutant, killed by at least one test, is killed only by tests that also kill it \citep{offutt2014}. Stated on the partition lattice, the
criterion covers monitors and interpretability probes as well as tests, and \S\ref{sub:deception} uses the same
join.

\paragraph{On \S\ref{sub:deception}.} Proposition~\ref{prop:deception} gives the structural form of a concern already raised operationally. Control protocols are built
on the assumption that some component is untrusted and must be checked by something it does not
control \citep{aicontrol2023}. Safety cases for untrusted monitoring turn on whether monitor and
monitored can collude \citep{untrusted2026}; collusion is, in the present terms, the loss of a
channel from the uncontrolled join. Chain-of-thought monitorability has been described as a
fragile opportunity \citep{cotmonitor2025}; the fragility here has a source, since a reasoning trace
the model emits is a channel it controls.

\begin{figure}[htbp]
\centering
\begin{tikzpicture}[scale=0.88, every node/.style={font=\scriptsize}]
\newcommand{\bx}{1.5}
\newcommand{\dbox}[2]{\draw[very thick] (#1,#2) rectangle (#1+\bx,#2+\bx);}
\newcommand{\dhor}[3]{\draw[very thick] (#1,#2+#3*\bx)--(#1+\bx,#2+#3*\bx);}
\newcommand{\dver}[3]{\draw[very thick] (#1+#3*\bx,#2)--(#1+#3*\bx,#2+\bx);}

%%%%%%%%%% row 1: the uncontrolled channels, joined
\def\ry{2.95}
\dbox{0}{\ry} \dhor{0}{\ry}{0.5}
\node[anchor=north] at (0.75,\ry-0.12) {$\parti{i_1}$};
\node at (1.85,\ry+0.75) {$\vee$};
\dbox{2.2}{\ry} \dver{2.2}{\ry}{0.667}
\node[anchor=north] at (2.95,\ry-0.12) {$\parti{i_2}$};
\node at (4.05,\ry+0.75) {$\vee$};
\node at (4.45,\ry+0.75) {$\cdots$};
\node at (4.85,\ry+0.75) {$\vee$};
\dbox{5.2}{\ry} \dhor{5.2}{\ry}{0.25}
\node[anchor=north] at (5.95,\ry-0.12) {$\parti{i_{k-1}}$};
\node at (7.05,\ry+0.75) {$=$};
\dbox{7.4}{\ry} \dhor{7.4}{\ry}{0.5} \dhor{7.4}{\ry}{0.25} \dver{7.4}{\ry}{0.667}
\node[anchor=north, align=center] at (8.15,\ry-0.12)
 {$\bigvee_{i\neq j}\parti{i}$\\(uncontrolled)};
\node[anchor=west, font=\scriptsize\itshape] at (9.25,\ry+0.75)
 {every uncontrolled channel};
\node[anchor=west, font=\scriptsize\itshape] at (9.25,\ry+0.40)
 {contributes its own lines};

%%%%%%%%%% row 2: overlay the controlled channel onto that join
\def\qy{0}
\dbox{0}{\qy} \dver{0}{\qy}{0.333} \dver{0}{\qy}{0.667}
\node[anchor=north, align=center] at (0.75,\qy-0.12) {$\parti{j}$\\(controlled)};
\node at (1.85,\qy+0.75) {$\vee$};
\dbox{2.2}{\qy} \dhor{2.2}{\qy}{0.5} \dhor{2.2}{\qy}{0.25} \dver{2.2}{\qy}{0.667}
\node[anchor=north, align=center] at (2.95,\qy-0.12) {$\bigvee_{i\neq j}\parti{i}$\\(top row)};
\node at (4.05,\qy+0.75) {$=$};
\dbox{4.4}{\qy} \dhor{4.4}{\qy}{0.5} \dhor{4.4}{\qy}{0.25} \dver{4.4}{\qy}{0.667}
\draw[line width=2.8pt, gray!60] (4.4+0.333*\bx,\qy)--(4.4+0.333*\bx,\qy+\bx);
\dver{4.4}{\qy}{0.333}
\node[anchor=south, font=\scriptsize\bfseries] at (4.4+0.333*\bx,\qy+\bx+0.06) {hideable};
\node[anchor=north, align=center] at (5.15,\qy-0.12) {a line only $\parti{j}$\\draws};
\node[anchor=west, font=\scriptsize\itshape, text width=4.3cm] at (6.35,\qy+0.75)
 {the other line of $\parti{j}$ is reproduced by an uncontrolled channel, so falsifying it
 creates a visible inconsistency};
\end{tikzpicture}
\caption{What a controlled channel can hide (\S\ref{sub:deception}). \textbf{Top:} the $k-1$ channels $B$ does not
control are joined, each contributing whatever lines it separates; this is the same overlay
operation as Figure~\ref{fig:grids}(c). \textbf{Bottom:} the controlled channel
$\parti{j}$ is laid over that join. Its right-hand line is corroborated, since an uncontrolled channel draws it too, so $B$ cannot move $A$'s belief there without inconsistency. Its
left-hand line appears nowhere else, so no other channel contradicts $B$ along that distinction, and by Proposition~\ref{prop:deception} $B$ can falsify it without failing the consistency check. The hideable part is a \emph{set of distinctions}, not a region.}
\label{fig:deception}
\end{figure}
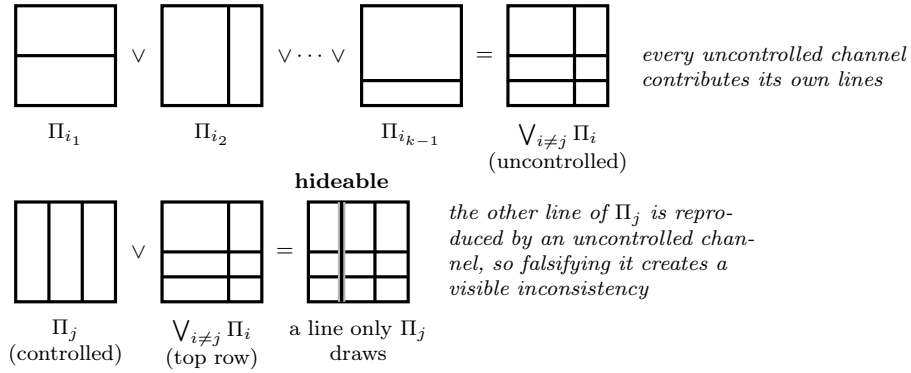

\section{Proofs and details for Section~\ref{sec:predictions}}\label{app:s5}

\begin{proof}[Proof of Proposition~\ref{prop:noise}]
From $\sigma_e$ on, the trigger or its link is gone, and the last clause of the assumption gives
the replies of the unconcealed version. Where the trigger is in the input and its distinction is the part that broke, an item with it and the same item without it now share a cell, and by Lemma~\ref{lem:bound} $B$ replies alike to them.
Below $\sigma_e$ the trigger and its link hold, and the clause on parts below their breaking point
gives the rest. The third claim is Corollary~\ref{cor:necessity} applied to the noised examinee,
and the last holds because an examinee that conceals nothing has no trigger or link.
\end{proof}

\section{Further related work}\label{app:related}

\paragraph{Ordering information structures.} Ranking observers by what they can tell apart
predates the AI setting. \citet{blackwell1953} orders statistical experiments by garbling: one
experiment is at least as informative as another for every decision problem exactly when the
second is the first composed with a noise kernel, so a finer structure can reproduce a coarser one
and not conversely. Theorems~\ref{thm:emulate} and~\ref{thm:detect} give that one-wayness on partitions, with the kernel
replaced by the coarsening map $h$. In epistemic logic the join of several agents' partitions is distributed knowledge \citep{fagin1995} and their meet is common knowledge \citep{aumann1976};
Proposition~\ref{prop:join} is a statement about the join, read as the discrimination an ensemble
of monitors has between them.

\paragraph{Relation to the classical material.} The proofs use three standard tools: a partition for behavioural distinguishability, the refinement order on partitions, and the factorisation of a map through a coarser quotient. Under Axiom~\ref{ax:noshortcut} the results bind any system whose behaviour is mediated by its percept, whatever its architecture. Simulation preorders \citep{park1981} state when one system can mimic
another, and testing equivalences \citep{denicola1984} equate processes no observer can
separate. The resulting bound on a \emph{test} is one-directional: the asymmetry belongs to the
evaluator--examinee pair rather than to either agent, which is why eliciting the examinee harder
cannot relieve it. The consequences of Section~\ref{sec:cons} follow from the same model, and in the
monitor case it gives an exact criterion (Proposition~\ref{prop:join}) where the findings on juries of LLM judges supply a tendency.

\paragraph{Renaming as gauge.} That latent representations are determined only up to a symmetry
is standard in identifiability theory \citep{comon1994,hyvarinen1999}. A renaming within cells (Proposition~\ref{prop:rename}) is invisible in the way an ICA permutation is, to any observer reading only which cell was occupied, though unlike a permutation it need not be invertible.

\paragraph{Hiding under partial observation.} Opacity asks whether a predicate on a system's runs
can ever be established by an observer restricted to a coarser view of them \citep{bryans2008}.
Definition~\ref{def:blind} asks an analogous, distributional question for a substitution rather than a predicate, and
Proposition~\ref{prop:blind} answers it: the substitution is invisible exactly when it leaves the
observer's cell frequencies unchanged.

\section{Notation}\label{app:notation}

\begin{table}[h]
\centering\small
\begin{tabular}{@{}ll@{}}
\toprule
Symbol & Meaning \\
\midrule
$\Sti$, $s$ & stimuli, and a single stimulus \\
$\mathcal{A}$ & actions an agent can emit; $\mathcal{A} \subseteq \Sti$ for the echo test (Definition~\ref{def:echo}) \\
$A, B, C$ & agents; in the statements, $A$ is the evaluator and $B$ the examinee \\
$P_Z : \Sti \to X_Z$ & perception of agent $Z$, with percept space $X_Z$ \\
$E_Z : X_Z \to \mathcal{A}$ & emission of agent $Z$ \\
$\beta_Z = E_Z \circ P_Z$ & behaviour of agent $Z$ (Axiom~\ref{ax:noshortcut}) \\
$\equiv_Z$, $\parti{Z}$ & equivalence induced by $P_Z$, and its discrimination partition \\
$\parti{B} \finer \parti{A}$ & $\parti{B}$ refines $\parti{A}$: $B$ is at least as sharp as $A$ \\
$\vee$, $\bigvee$ & join: the coarsest partition refining its arguments (Definition~\ref{def:join}) \\
$h$ & coarsening map $P_B(\Sti) \to P_A(\Sti)$, with $P_A = h \circ P_B$ (Theorem~\ref{thm:emulate}) \\
$\rho$ & prototype selector $P_A(\Sti) \to \mathcal{A}$, with $P_A \circ \rho = \mathrm{id}$ \\
$\mathcal{T} = (S, G)$, $Q$ & a test: its cases and acceptance rule; the cases probed (Definition~\ref{def:test}) \\
$G(s)$ & acceptable replies to $s$ under a test (Definition~\ref{def:test}) \\
$K$ & a cell of the partition in play \\
$O_i$ & observers whose collective discrimination is a join (\S\ref{sub:join}) \\
$J$ & a percept-reading evaluator (Figure~\ref{fig:access}) \\
$p$, $\phi$ & an interpretability probe, and the representation it reads (\S\ref{sub:absence}) \\
$o_i$, $c_i$, $\parti{i}$ & reading of channel $c_i$, and its partition (Definition~\ref{def:channel}) \\
$U$ & join of the channels $B$ does not control \\
$q$, $T$ & map from a stimulus to its $A$-cell; a covert encoding \\
$\mu$, $f_{*}\mu$ & distribution on stimuli; its pushforward along $f$ \\
$d$, $V_d$ & a detector, and the cells on which it fires \\
$\mathrm{TPR}(d)$, $\mathrm{FPR}(d)$ & true and false positive rates of a detector (\S\ref{sub:sandbag}) \\
$\sigma$, $\sigma_e$ & noise scale; the lower breaking point of the trigger and its link (Proposition~\ref{prop:noise}) \\
\bottomrule
\end{tabular}
\caption{Notation.}
\label{tab:notation}
\end{table}

\end{document}